\documentclass{article}

\PassOptionsToPackage{numbers, compress}{natbib}
 \usepackage[final]{nips_2016}

\usepackage[utf8]{inputenc} % allow utf-8 input
\usepackage[T1]{fontenc}    % use 8-bit T1 fonts
\usepackage{url}            % simple URL typesetting
\usepackage{booktabs}       % professional-quality tables
\usepackage{amsfonts}       % blackboard math symbols
\usepackage{nicefrac}       % compact symbols for 1/2, etc.
\usepackage{microtype}      % microtypography
\usepackage{amssymb}
\usepackage{epsfig}
\usepackage{amsfonts}
\usepackage{mathrsfs}
\usepackage{amsmath}
\usepackage{graphicx}
\usepackage{color}
\usepackage{caption}
\usepackage{picinpar}
\usepackage{booktabs}
\usepackage{float}
\usepackage{url}
\usepackage{dsfont}
\usepackage{multirow}
\usepackage{subfloat}
\usepackage{slashbox}
\usepackage{array}
\usepackage{subfigure}
\usepackage[none]{hyphenat}
\title{Stochastic Configuration Networks:\\ Fundamentals and  Algorithms}

\author{
  Dianhui~Wang\thanks{Corresponding author.}, ~~Ming Li \\ \\
  Department of Computer Science and Information Technology\\ 
  La Trobe University, 
  Melbourne, VIC 3086, Australia \\
  \texttt{Email:dh.wang@latrobe.edu.au} \\
}

\begin{document}

\maketitle

\begin{abstract}
This paper contributes to a development of randomized methods for neural networks. The proposed learner model is generated incrementally by stochastic configuration (SC) algorithms, termed as Stochastic Configuration Networks (SCNs). In contrast to the  existing randomised learning algorithms for single layer feed-forward neural networks (SLFNNs), we randomly assign the input weights and biases of the hidden nodes in the light of a supervisory mechanism, and the output weights are  analytically evaluated in either constructive or selective manner. As fundamentals of SCN-based data modelling techniques, we establish some theoretical results on the universal approximation property. Three versions of SC algorithms are presented for  regression problems (applicable for classification problems as well) in this work. Simulation results concerning  both function approximation and real world data regression indicate some remarkable merits of our proposed SCNs in terms of less human intervention on the network size setting, the scope adaptation of random parameters, fast learning and sound generalization.
\end{abstract}
\section{Introduction}
Over the past decades, neural networks have been successfully applied for data modelling due to its universal approximation power for nonlinear maps \cite{Chen1995,Cybenko1989,Hornik1989,Park1991}, and learning capability from a collection of training samples \cite{Hecht-Nielsen1988,Rumelhart1986}. In practice, however, it is quite challenging to properly determine an appropriate architecture (here refers to the number of hidden nodes) of a neural network so that the resulting learner model can achieve sound performance for both learning and generalization. To resolve this problem, one turns to develop  constructive approaches for building neural networks, starting with a small sized network, followed by incrementally generating hidden nodes (corresponding to a set of input weights and biases) and output weights until a pre-defined termination criterion meets. Being a fundamental of manifold applications, it is essential to ensure that the constructive neural network shares the universal approximation property. In  \cite{Barron1993}, Barron  proposed a greedy learning framework based on the work reported in \cite{Jones1992}, and established some significant results on the convergence rate. In \cite{Kwok1997}, Kwok and Yeung presented a method to construct neural networks through optimizing some objective functions. Theoretically, their proposed   algorithm can generate universal approximators for any continuous  nonlinear functions provided that the activation function meets some conditions. It has been aware that the process of iterative searching for an appropriate set of parameters (input weights, biases and output weights) is time-consuming and computationally intensive although  the universal approximation property holds, but very difficult to be employed for dealing with large-scale data analytics.  

Randomized approaches for large-scale computing are highly desirable due to their effectiveness and efficiency \cite{Mahoney2011}. In machine learning for data modelling, randomized algorithms have demonstrated great potential in developing fast learner models and learning algorithms with much less computational cost \cite{LukoandJaeger2009,ScardapaneandWang2017,SI-ProfWang2016}. Readers are strongly recommended to refer to our survey paper \cite{ScardapaneandWang2017} for more details.  From algorithm perspectives, randomized learning techniques for neural networks received attention in later 80's \cite{Broomhead1988} and further developed in early 90's \cite{Pao1994,Pao1992,Schmidt1992}. A common and basic idea behind these randomized learning algorithms is a two-step training paradigm, that is, randomly assigning the input weights and biases of neural networks and evaluating the output weights by solving a linear equation system using the well-known least squares method and its regularized versions.  From approximation theory viewpoint, randomized Radial Basis Function (RBF) Networks proposed in \cite{Broomhead1988} and Random Vector Functional-link (RVFL) networks proposed in \cite{Pao1992} can be regarded as Random Basis Approximators (RBAs) \cite{Tyukin2009}. Thus, it is essential and interesting to look into the universal approximation capability of RBAs in the sense of probability. In \cite{Igelnik1995}, Igelnik and Pao proved that a RVFL network with random parameters (input weights and biases) chosen from the  uniform distribution defined over a range can be a universal approximator with probability one for continuous functions. In \cite{Husmeier1999}, Husmeier revisited this significant result and showed that the universal approximation property of RVFL networks holds as well for symmetric interval setting of the random parameter scope if the function to be approximated meets Lipschitz condition. In \cite{Tyukin2009}, Ivan et al.  empirically investigated the feasibility of RBAs for data modelling and showed that a supervisory mechanism is necessary to make RVFL networks applicable. Their experiments clearly indicate that RVFL networks  fail to approximate a target function with a very high probability if the setting for random parameters is improper. This phenomenon was further studied with mathematical justifications in \cite{SI-Gorban2016}. From implementation considerations, we need to know two key parameters involved in design of  RVFL networks: the number of hidden nodes and the scope of random parameters. Indeed, as one of the special features of this type of learner models, the first parameter associated with the modelling accuracy must be set largely. The second parameter is related to the approximation capability of the class of random basis functions.  Obviously, these settings play very  important roles to successfully build randomized learner models for real world applications. Intuitively, constructive or incremental RVFL (IRVFL) networks may be a possible solution for problem solving. However, our recent work reported in \cite{LiandWang2016} reveals the infeasibility of IRVFL networks, if they are incrementally built  with random input weights and biases assigned in a fixed scope and its convergence rate satisfies some certain conditions. Thus, further researches on supervisory mechanisms with adaptive scope setting in random assignment of the input weights and biases to ensure the  universal approximation capability of IRVFL networks are being expected.  

To  the development of randomized learning techniques for neural networks, our prime and original contribution from this paper  is the way of assigning the random parameters with an inequality constraint and adaptively selecting the scope of the random parameters, ensuring the universal approximation property of the built randomized learner models. Indeed, this work firstly touches the base of implementation of the random basis function approximation theory. It should be clarified that one cannot view SCNs as a specific implementation of RVFL networks due to some remarkable distinctions in randomization of the learner model.  Three algorithmic implementations of SCNs, namely Algorithm SC-I, SC-II and SC-III, are presented with the same supervisory mechanism for  configuring the random parameters, but different  methods for computing the output weights. Concretely, SC-I employs a constructive scheme to evaluate the output weights only for the newly added hidden node and keep  all of the previously obtained output weights unchanged; SC-II recalculates a part of the current output weights by solving a local least squares problem with a user specified shifting window size; and SC-III finds the output weights all together through  solving a global least squares problem with the current learner model. Our experimental results on a toy example for function approximation and real world regression tasks demonstrate remarkable improvements on modelling performance, compared with the results obtained by existing methods such as Modified Quickprop (MQ) \cite{Kwok1997} and IRVFL \cite{LiandWang2016}. 

The remainder of this paper is organized as follows: Section 2 briefly reviews constructive neural networks with a recall for the infeasibility of IRVFL networks. Section 3 details our proposed stochastic configuration networks with both theoretical analysis and algorithmic description. Section 4 reports our simulation results, and Section 5  concludes this paper with some remarks on further studies. 

The following notation is  used throughout this paper. Let $\Gamma:=\{g_1, g_2, g_3...\}$ be a set of real-valued functions, span$(\Gamma)$ denote a function space spanned by $\Gamma$; $L_{2}(D)$ denote the space of all Lebesgue measurable functions $f=[f_1,f_2,\ldots,f_m]:\mathbb{R}^{d}\rightarrow \mathbb{R}^{m}$ defined on $D\subset \mathbb{R}^{d}$, with the $L_2$ norm defined as
\begin{equation}\label{multiple_lp}
  \|f\|:=\left(\sum_{q=1}^{m}\int_{D}|f_q(x)|^2dx\right)^{1/2}<\infty.
\end{equation}
The inner product of $\theta=[\theta_1,\theta_2,\ldots,\theta_m]:\mathbb{R}^{d}\rightarrow \mathbb{R}^{m}$ and $f$ is defined as
\begin{equation}\label{multiple_inner}
  \langle f,\theta\rangle:=\sum_{q=1}^{m}\langle f_q,\theta_q\rangle=\sum_{q=1}^{m}\int_{D}f_q(x)\theta_q(x)dx.
\end{equation}
In the special case that $m=1$, for a real-valued function $\psi:\mathbb{R}^{d}\rightarrow \mathbb{R}$ defined on $D\subset \mathbb{R}^{d}$, its $L_2$ norm becomes $ \|\psi\|:=(\int_{D}|\psi(x)|^2dx)^{1/2}$, while the inner product of $\psi_1$ and $\psi_2$ becomes $\langle \psi_1,\psi_2\rangle=\int_{D}\psi_1(x)\psi_2(x)dx$.

\section{Related Work}
Instead of training a learner model with a fixed architecture, the process of constructive neural networks starts with a small sized network then adds hidden nodes incrementally until an acceptable tolerance is achieved. This approach does not require any prior knowledge about the complexity of the network for a given task. This section briefly reviews some closely related work on constructive neural networks. Some comments on these methods are also given. 

Given a target function $f:\mathds{R}^{d}\rightarrow \mathds{R}$, suppose that we have already built a SLFNN with $L-1$ hidden nodes, i.e., $f_{L-1}(x)=\sum_{j=1}^{L-1}\beta_jg_j(w_j^\mathrm{T}x+b_j)$ ($L=1,2,\ldots$, $f_0=0$), and the current residual error, denoted as $e_{L-1}=f-f_{L-1}$, does not reach an acceptable tolerance level. The construction process is concerned with how to incrementally add $\beta_L$, $g_L$ ($w_L$ and $b_L$) leading to $f_{L}=f_{L-1}+\beta_Lg_L$ until the residual error falls into an expected tolerance $\epsilon$.
\subsection{Constructive Neural Networks: Deterministic Methods}
%\subsection{Deterministic Methods}
%-----------------------------------------------------------------------------------------------------------------------------------------------------%
%% Review of Barron's work in 1993
%-----------------------------------------------------------------------------------------------------------------------------------------------------%
In  \cite{Barron1993}, Barron proposed a greedy approximation framework based on Jones' Lemma \cite{Jones1992}. The main result can be stated in the following theorem. 

\textbf{Theorem 1 (Barron \cite{Barron1993}).} Given a Hilbert space $G$, suppose $f\in \overline{\mbox{conv}(G)}$ (closure of the convex hull of the set G), and $\|g\|\leq b_g$ for each $g \in G$. Let $b_f>c_f>\sup \|g\|+\|f\|$. For $L=1, 2, \ldots$, the sequence of approximations $\{f_L\}$ is deterministic and described as
\begin{equation}\label{barron_th}
  f_L=\alpha_Lf_{L-1}+(1-\alpha_L)g_L,
\end{equation}
where
\begin{equation}
  \alpha_L=\frac{b_f^2}{b_f^2+\|f_{L-1}-f\|^2},
\end{equation}
and $g_L$ is chosen such that the following condition holds
\begin{equation}\label{greedy_condition}
  \langle f_L-f,g_L-f\rangle<\frac{b_f^2-c_f^2}{2b_f^2}\|f_{L-1}-f\|^2.
\end{equation}
Then, for every $L\geq1$, $\|f-f_L\|^2\leq \frac{C^{'}}{L}$, where $C^{'}\geq b_f$.

\textbf{Remark 1.} Some optimization techniques are needed in order to find a suitable $g_L$ to meet the condition (\ref{greedy_condition}). In fact, the whole process aims to  minimize $\|\alpha_Lf_{L-1}+(1-\alpha_L)g_{L}-f\|$ at each iteration, which is functionally equivalent to choosing $g_L\in G$ that maximizes $\langle f-f_{L-1},g\rangle$. Note that this constructive scheme is only applicable to the target function $f$ belonging to the closure of the convex hull of $G$, that means the convergence to all $L_2$ functions cannot be guaranteed under this framework. In addition, the new leaner model $f_L$ comes from a specific convex combination of the previous model $f_{L-1}$ and the newly added term $g_L$, which results in a weak solution compared against one obtained through optimizing the left hand side of (\ref{greedy_condition}) with respect to the coefficient $\alpha_L$.

%-----------------------------------------------------------------------------------------------------------------------------------------------------%
%%%%%%Review of Kwok97's work %%%%%%%%%%%%%%%%%%%%%%%%%%%%%%%%%%%%%%%%%%%%%%%%%%%%%%%%%%%%%%%%%%%%%%%%%%%%%%%%%%%%%%%%%%%%%%%%%%%%%%%%%%%%%%%%%%%%%%%%%
%-----------------------------------------------------------------------------------------------------------------------------------------------------%
In \cite{Kwok1997}, Kwok and Yeung proposed an incremental learning strategy for building SLFNNs and proved the universal approximation property. At each step, the input weights and biases ($w$ and $b$) of the newly added node at the hidden layer are obtained by maximizing an objective function and the output weights are evaluated using the least squares method. This theoretical result can be stated as follows:

\textbf{Theorem 2 (Kwok and Yeung \cite{Kwok1997}).} Let $\Gamma$ be a set of basis function. Assume that span($\Gamma$) is dense in $L_2$ space and $\forall g\in \Gamma$, $0<\|g\|<b_g$ for some $b_g\in \mathds{R}^{+}$. If $g_L$ is selected as to maximize $\langle e_{L-1},g_L\rangle^2/\|g\|^2$, then $\lim_{L\rightarrow +\infty}\|f-f_L\|=0$.

\textbf{Remark 2.} Theoretically, the convergence property can be guaranteed provided that we can find $g_L$ to maximize $\langle e_{L-1},g_L\rangle^2/\|g\|^2$. In practice, however, local minima occurs frequently when performing the gradient ascent algorithm. Sometimes, during the course of incremental learning, the optimization process in constructing a new hidden node seems meaningless as the updating of hidden parameters is excessively slow, which means the reduction of residual error will be close to zero. In fact, this point was discussed in \cite{Kwok1994} and a Modified Quickprop algorithm was suggested in \cite{Kwok1997} in order to alleviate this problem. However, this weakness cannot be overcome completely for some complex tasks. In essence, the Modified Quickprop algorithm that iteratively finds the appropriate hidden parameters ($w$ and $b$) may still face some obstacles in the optimization process of the objective function when it is searching in a plateau of the error surface. Once $w$ ($b$) at some iterative step falls in a region that both the first and second derivatives of the objective function with respect to $w$ ($b$) are nearly zero, the learning reaches a halt and may be terminated. In a nut shell, the derivative-based optimization processes suffer from some inherent shortcomings and have lower probability to generate a universal learner eventually. 

The constructive methods mentioned above consist of two phases: selecting the hidden parameters (the input  weights and biases) according to some certain criteria and evaluating the output weights after adding a new hidden node. Although the universal approximation property can be guaranteed by incrementally adding the hidden nodes,  the resulting SLFNNs usually need a quite few hidden nodes to achieve good learning performance. In practice, seeking for a basis function through maximizing $\langle e_{L-1},g_L\rangle^2/\|g\|^2$ is very time consuming. Thus, for many real world applications, deterministic methods used in building constructive neural networks seem have no or less applicability. As one of pathways to generate neural networks with the universal approximation property (corresponding to the perfect learning power), randomized approaches have great potential to access a faster and feasible solution \cite{LukoandJaeger2009,ScardapaneandWang2017,SI-ProfWang2016}.
\subsection{Constructive Random Basis Approximators}
Random vector functional-link (RVFL) networks \cite{Pao1994, Pao1992} can be regarded as a randomized version of SLFNNs, where the input weights and biases are randomly assigned and fixed during the training phase, and the output weights are analytically evaluated by the least squares method \cite{Lancaster1985}.  In \cite{Igelnik1995}, Igelnik and Pao theoretically justified its universal approximation property based on Monte-Carlo method with  the limit-integral representation of the target function. The main result can be restated as follows:  

\textbf{Theorem 3 (Igelnik and Pao \cite{Igelnik1995}).} For any compact set $D\subset\mathds{R}^d$, given $f\in C(D)$ (i.e., the set of all continuous functions defined over $D$), and any activation function $g$ that satisfies $\int_{\mathds{R}}|g(t)|^2 dt<\infty$ or $\int_{\mathds{R}}|g^{'}(t)|^2dt<\infty$, there exist a sufficiently large $L$, a set of $\beta_1,\beta_2,...,\beta_L $ and a probabilistic space $\chi_L$, such that $f_L=\sum_{j=1}^L\beta_jg(x;w_j,b_j)$ can approximate $f$ with arbitrary accuracy in the sense of probability one, if $w$ and $b$ are randomly assigned over $\chi_L$ and follow certain distribution. Alternatively, this result can be expressed as 
\begin{equation}
\lim_{L\rightarrow+\infty}E\Big(\int_{D}|f(x)-f_{L}(x)|^2dx\Big)=0,
\end{equation}
where $E$ is the expectation operator with respect to the probabilistic space $\chi_L$.

\textbf{Remark 3.}   In \cite{Husmeier1999}, Husmeier revisited the universal approximation property of RVFL networks with a symmetric interval setting for the random parameters. Strictly speaking, such a property holds for only target functions satisfying Lipschitz condition. From our experience, however, most of data modelling  problems from real world applications meet Lipschitz condition. Thus, for simplicity, we adopt the symmetric interval setting for the random parameters in this paper.  It should be aware that such a special setting does not limit the development of our framework at all, and it is just a matter of implementation indeed. 

The theoretical result stated in Theorem 3 is fundamental and significant for building randomized  neural networks. Similar to the case of making use of the SLFNNs in resolving real world problems, we need to develop effective algorithms to implement such a class of randomized predictive  models. It is natural to think of an  incremental implementation of RVFL (IRVFL) networks, where the model is built incrementally with random assignment of the input weights and biases, and constructive evaluation of its output weights. Although the construction process of IRVFL networks seems to be computationally efficient, unfortunately the universal approximation property of the constructed learner cannot be guaranteed. The following Theorem 4 from our recent work \cite{LiandWang2016} have justified this point.

\textbf{Theorem 4 (Li and Wang \cite{LiandWang2016}).} Let span($\Gamma$) be dense in $L_2$ space and $\forall g\in \Gamma$, $0<\|g\|<b_g$ for some $b_g\in \mathds{R}^{+}$.
Suppose that $g_L$ is randomly generated and $\beta_L$ is given by
\begin{equation}\label{th4_con1}
\beta_{L}=\frac{\langle e_{L-1},g_L\rangle}{\|g_L\|^2}.
\end{equation}
For sufficiently large $L$, if the followings hold:
\begin{eqnarray}\label{th4_con2}
 \frac{\|e_{L-1}\|^2-\|e_{L}\|^2}{\|e_{L-1}\|^2}\leq\varepsilon_L<1,
\end{eqnarray}
and
\begin{eqnarray}\label{th4_con3}
 \lim_{L\rightarrow \infty}\prod_{k=1}^{L}(1-\varepsilon_k)=\varepsilon>0.
\end{eqnarray}
Then, the constructive neural network with random weights, $f_L$, has no universal approximation capability, that is,
\begin{eqnarray}
\lim_{L\rightarrow \infty}\|f-f_L\|\geq\sqrt{\varepsilon}\|f\|.
\end{eqnarray}
Theorem 4 reveals that IRVFL networks may not share the universal approximation property, if the output weights are taken as (\ref{th4_con1}) and the residual error sequence meets conditions (\ref{th4_con2}) and (\ref{th4_con3}). The consequence stated in Theorem 4 still holds if the output weights are evaluated using the least squares method. We state this interesting result in the following Theorem 5. 

\textbf{Theorem 5.} Let span($\Gamma$) be dense in $L_2$ space and $\forall g\in \Gamma$, $0<\|g\|<b_g$ for some $b_g\in \mathds{R}^{+}$.
Suppose that $g_L$ is randomly generated and the output weights are calculated by solving the global least square problem, i.e.,
$[\beta_1^{*}, \beta_2^{*},\ldots,\beta_{L}^{*}]=\arg \min_{\beta}\|f-\sum_{j=1}^{L}\beta_jg_j\|$. 
For sufficiently large $L$, if (\ref{th4_con2}) and (\ref{th4_con3}) hold for the residual error sequence $\|e_L^{*}\|$ (corresponding to $\|e_L\|$ in Theorem 4), the constructive neural network with random hidden nodes has no universal approximation capability, that is,
\begin{eqnarray}\label{th5_con}
\lim_{L\rightarrow \infty}\|f-\sum_{j=1}^{L}\beta_j^{*}g_j\|\geq\sqrt{\varepsilon}\|f\|.
\end{eqnarray}
\textbf{Proof.} Simple computations can verify that the sequence $\|e_{L}^{*}\|^2$ is monotonically decreasing and converges. Indeed, let $\tilde{\beta}_{L}=\langle e_{L-1}^{*},g_L\rangle/\|g_L\|^2$, we have 
\begin{eqnarray}
\|e_{L}^{*}\|^2&\leq&\langle e_{L-1}^{*}-\tilde{\beta}_{L}g_L,e_{L-1}^{*}-\tilde{\beta}_{L}g_L\rangle \nonumber\\ 
&=&\|e_{L-1}^{*}\|^2-\frac{\langle e_{L-1}^{*},g_L\rangle^2}{\|g_L\|^2}\nonumber\\
&\le& \|e_{L-1}^{*}\|^2.
\end{eqnarray} 
Then, (\ref{th5_con}) can be easily obtained by following the proof of Theorem 4.

Clearly, the universal approximation property is conditional to IRVFL networks whatever the output weights are evaluated. This happens due to multiple reasons (e.g. the scope setting and/or the improper way to assign the random parameters) that is hard to tell mathematically. In this paper, we propose a solution for constructing randomized learner models under a supervisory mechanism to ensure the universal approximation property.  

\section{Stochastic Configuration Networks}
Universal approximation property is fundamental to a learner model for data modelling. Logically, one cannot expect to build a neural network with good generalization but poor learning performance. Therefore, it is essential to share the universal approximation property for SCNs. This section details our proposed SCNs, including  proofs of the universal approximation property and algorithm descriptions. Some remarks on algorithmic implementations are also given.
\subsection{Universal Approximation Property}
Given a target function $f:\mathds{R}^{d}\rightarrow \mathds{R}^{m}$, suppose that a SCN with $L-1$ hidden nodes has already been constructed, that is, $f_{L-1}(x)=\sum_{j=1}^{L-1}\beta_jg_j(w_j^\mathrm{T}x+b_j)$ ($L=1,2,\ldots$, $f_0=0$), where $\beta_j=[\beta_{j,1},\ldots,\beta_{j,m}]^\mathrm{T}$. Denoted the current residual error by $e_{L-1}=f-f_{L-1}=[e_{L-1,1},\ldots,e_{L-1,m}]$. If $\|e_{L-1}\|$ does not reach a pre-defined tolerance level, we need to generate a new random basis function $g_L$ ($w_L$ and $b_L$) and evaluate the output weights $\beta_L$ so that the leading model $f_{L}=f_{L-1}+\beta_Lg_L$ will have an improved residual error.  In this paper, we propose a method to randomly assign the input weights and biases with a supervisory mechanism (inequality constraint), and provide with three ways to evaluate the output weights of SCNs with $L$ hidden nodes (i.e., after adding the new node).  Mathematically, we can prove that the resulting randomized learner models incrementally built based on our stochastic configuration idea are universal approximators.  

\textbf{Theorem 6.} Suppose that  span($\Gamma$) is dense in $L_2$ space and $\forall g\in \Gamma$, $0<\|g\|<b_g$ for some $b_g\in \mathds{R}^{+}$. Given $0<r<1$ and a nonnegative real number sequence $\{\mu_L\}$ with $\lim_{L\rightarrow+\infty}\mu_L=0$ and $\mu_L\leq (1-r)$. For $L=1,2,\ldots$, denoted by 
\begin{equation}\label{delta1}
\delta_{L}=\sum_{q=1}^{m}\delta_{L,q}, \delta_{L,q}=(1-r-\mu_L)\|e_{L-1,q}\|^2, q=1,2,...,m.
\end{equation}
If the random basis function $g_L$ is generated to satisfy the following inequalities:
\begin{equation}\label{step1}
%\sum_{q=1}^m\langle e_{L-1,q},g_L\rangle^2\geq b_g^2\delta_L,
\langle e_{L-1,q},g_L\rangle^2\geq b_g^2\delta_{L,q}, q=1,2,...,m,
\end{equation}
and the output weights are constructively evaluated by
\begin{equation}\label{step2}
\beta_{L,q}=\frac{\langle e_{L-1,q},g_L\rangle}{\|g_L\|^2}, q=1,2,\ldots,m.
\end{equation}
Then, we have $\lim_{L\rightarrow +\infty}\|f-f_L\|=0,$ where $f_L=\sum_{j=1}^{L}\beta_{j}g_j$, $\beta_j=[\beta_{j,1},\ldots,\beta_{j,m}]^{\mathrm{T}}$.

\textbf{Proof.} According to (\ref{step2}), it is easy to verify that $\{\|e_{L}^2\|\}$ is monotonically decreasing. Thus, the sequence $\|e_L\|$ is convergent as $L\rightarrow\infty$. From (\ref{delta1}), (\ref{step1}) and (\ref{step2}), we have 
\begin{eqnarray}
&&\|e_L\|^2-(r+\mu_L)\|e_{L-1}\|^2\nonumber\\&=& \sum_{q=1}^m\left(\langle e_{L-1,q}-\beta_{L,q}g_L,e_{L-1,q}-\beta_{L,q}g_L\rangle-(r+\mu_L)\langle e_{L-1,q},e_{L-1,q}\rangle\right)\nonumber\\
&=&\sum_{q=1}^m\left((1-r-\mu_L)\langle e_{L-1,q},e_{L-1,q}\rangle-2\langle e_{L-1,q},\beta_{L,q}g_L\rangle+\langle\beta_{L,q}g_L,\beta_{L,q}g_L\rangle\right)\nonumber\\
&=&(1-r-\mu_L)\|e_{L-1}\|^2-\frac{\sum_{q=1}^m\langle e_{L-1,q},g_L\rangle^2}{\|g_L\|^2}\nonumber\\
&=&\delta_L-\frac{\sum_{q=1}^m\langle e_{L-1,q},g_L\rangle^2}{\|g_L\|^2}\nonumber\\
&\leq& \delta_L-\frac{\sum_{q=1}^m\langle e_{L-1,q},g_L\rangle^2}{b_g^2}\leq0.
\end{eqnarray}
Therefore,  the following inequality holds:
\begin{equation}\label{contract}
\|e_L\|^2\leq r\|e_{L-1}\|^2+\gamma_L, (\gamma_L=\mu_L\|e_{L-1}\|^2\geq 0).
\end{equation}
Note that $\lim_{L\rightarrow+\infty}\gamma_L=0$, by using (\ref{contract}), we can easily show that $\lim_{L\rightarrow +\infty}\|e_{L}\|^2=0$ which implies $\lim_{L\rightarrow +\infty}\|e_{L}\|=0$. This completes the proof of Theorem 6.\\

Theorem 6 provides a constructive scheme, i.e., (\ref{step1}) and (\ref{step2}), that can consequently lead to a universal approximator. Unlike the strategy that maximizes some objective functions in \cite{Kwok1997}, the supervisory mechanism (\ref{step1}), which aims at finding appropriate $w_L$ and $b_L$ for a new hidden node, weakens the demanding condition as required to achieve the maximum value for $\langle e_{L-1},g_L\rangle^2$ or $\langle e_{L-1},g_L\rangle^2/\|g_{L}\|^2$ (for the case $m=1$). In fact, the existence of $w_L$ and $b_L$ satisfying (\ref{step1}) can be easily deduced because $\Psi(w,b)=\sum_{q=1}^m\langle e_{L-1,q},g_L\rangle^2/\|g_{L}\|^2$ is a continuous function in the parameter space, and $\delta_L=(1-r-\mu_L)\|e_{L-1}\|^2$ can be far less than $\|e_{L-1}\|^2$ once the $r$ is selected to approach 1. Overall, the supervisory mechanism described in (\ref{step1}) not only makes it possible to randomly assign  the hidden parameters, which performs more flexibly and efficiently in generating a new hidden node, but also
enforces the residual error to be zero along with the constructive process.

\textbf{Remark 4.} Our proposed supervisory mechanism in (\ref{step1}) indicates  that random assignment of $w_L$ and $b_L$ should be constrained and data dependent.  That is to say, the configuration of hidden parameters needs to be relevant to the given training samples, rather than totally rely on the distribution and/or scoping parameter setting of the random weights and biases. To the best of our knowledge, our attempt on the supervisory mechanism  (\ref{step1}) is the first time in design of randomized learner models, and it fills the gaps between the scheme of solving a global nonlinear optimization problem (that are usually quite demanding in both time and space since iterative solution seems inevitable) and the strategy of freely random assignment of  the hidden parameters without any constraint. It is worth mentioning that Theorem 6 is still valid if the learning parameter $r$ is unfixed and set based on an increasing sequence approaching 1 (always less than 1). That makes SC algorithm be more flexible for the implementation of randomly searching $w_L$ and $b_L$ even when the residual error is smaller. It has been observed that finding out appropriate $w_L$ and $b_L$ for the newly added node becomes more challenging as the residual error becomes smaller. In this case, we need to set the value of $r$ to be extremely close to 1.

It is obvious that $\beta_L=[\beta_{L,1},\ldots,\beta_{L,m}]^{\mathrm{T}}$ in Theorem 6 is analytically evaluated by $\beta_{L,q}=\langle e_{L-1,q},g_L\rangle/\|g_L\|^2$ and remain fixed for further adding steps. This determination scheme, however, may cause very slow convergence rate for the constructive process. Thus, we consider a recalculation scheme for the output weights, that is, once $g_j (j=1,2,\ldots,L)$ have been generated according to (\ref{step1}), $\beta_1,\beta_2,\ldots,\beta_{L}$ can be evaluated by minimizing the global residual error. The following Theorem 7 gives a result on the universal approximation property if the least squares method is applied to update the output weights in a proceeding manner. 

Let $[\beta_1^{*}, \beta_2^{*},\ldots,\beta_{L}^{*}]=\arg \min_{\beta}\|f-\sum_{j=1}^{L}\beta_jg_j\|$, $e_{L}^{*}=f-\sum_{j=1}^{L}\beta_j^{*}g_j=[e_{L,1}^{*},\ldots,e_{L,m}^{*}]$, and define intermediate values $\tilde{\beta}_{L,q}=\langle e_{L-1,q}^{*},g_L\rangle/\|g_L\|^2$ for $q=1,\ldots,m$ and $\tilde{e}_{L}=e_{L-1}^{*}-\tilde{\beta}_{L}g_L$, where $\tilde{\beta}_{L}=[\tilde{\beta}_{L,1},\ldots,\tilde{\beta}_{L,m}]^{\mathrm{T}}$ and $e_0^{*}=f$.

\textbf{Theorem 7.} Suppose that  span($\Gamma$) is dense in $L_2$ space and $\forall g\in \Gamma$, $0<\|g\|<b_g$ for some $b_g\in \mathds{R}^{+}$. Given $0<r<1$ and a nonnegative real number sequence $\{\mu_L\}$ with $\lim_{L\rightarrow+\infty}\mu_L=0$ and $\mu_L\leq (1-r)$.  For $L=1,2,\ldots$, denoted by
\begin{equation}
%\delta_L^{*}=(1-r-\mu_L)\|e_{L-1}^{*}\|^2.
\delta_{L}^{*}=\sum_{q=1}^{m}\delta_{L,q}^{*}, \delta_{L,q}^{*}=(1-r-\mu_L)\|e_{L-1,q}^{*}\|^2, q=1,2,...,m.
\end{equation}
If the random basis function $g_L$ is generated to satisfy the following inequalities:
\begin{equation}\label{step3}
\langle e_{L-1,q}^{*},g_L\rangle^2\geq b_g^2\delta_{L,q}^{*}, q=1,2,...,m,
\end{equation}
and the output weights are evaluated by
\begin{equation}\label{step4}
[\beta_1^{*}, \beta_2^{*},\ldots,\beta_{L}^{*}]=\arg \min_{\beta}\|f-\sum_{j=1}^{L}\beta_jg_j\|.
\end{equation}
Then, we have $\lim_{L\rightarrow +\infty}\|f-f_L^{*}\|=0,$ where $f_L^{*}=\sum_{j=1}^{L}\beta_{j}^{*}g_j$, $\beta_{j}^{*}=[\beta^{*}_{j,1},\ldots,\beta^{*}_{j,m}]^{\mathrm{T}}$.

\textbf{Proof.} It is easy to show that $\|e_{L}^{*}\|^2\leq \|\tilde{e}_{L}\|^2=\|e_{L-1}^{*}-\tilde{\beta}_{L}g_L\|^2\leq \|e_{L-1}^{*}\|^2\leq \|\tilde{e}_{L-1}\|^2$ for $L=1,2,\ldots$, so $\{\|e_{L}^{*}\|^2\}$ is monotonically decreasing and convergent.
Hence, we have
\begin{eqnarray}
&&\|e_L^{*}\|^2-(r+\mu_L)\|e_{L-1}^{*}\|^2\nonumber\\&\leq&\|\tilde{e}_{L}\|^2-(r+\mu_L)\|e_{L-1}^{*}\|^2\nonumber\\
&=&\sum_{q=1}^{m}\left(\langle e_{L-1,q}^{*}-\tilde{\beta}_{L,q}g_L,e_{L-1,q}^{*}-\tilde{\beta}_{L,q}g_L\rangle-(r+\mu_L)\langle e_{L-1,q}^{*},e_{L-1,q}^{*}\rangle\right)\nonumber\\
&=&\sum_{q=1}^{m}\left((1-r-\mu_L)\langle e_{L-1,q}^{*},e_{L-1,q}^{*}\rangle-2\langle e_{L-1,q}^{*},\tilde{\beta}_{L,q}g_L\rangle+\langle\tilde{\beta}_{L,q}g_L,\tilde{\beta}_{L,q}g_L\rangle\right)\nonumber\\
&=&(1-r-\mu_L)\|e_{L-1}^{*}\|^2-\frac{\sum_{q=1}^{m}\langle e_{L-1,q}^{*},g_L\rangle^2}{\|g_L\|^2}\nonumber\\
&=&\delta_L^{*}-\frac{\sum_{q=1}^{m}\langle e_{L-1,q}^{*},g_L\rangle^2}{\|g_L\|^2}\nonumber\\
&\leq& \delta_L^{*}-\frac{\sum_{q=1}^{m}\langle e_{L-1,q}^{*},g_L\rangle^2}{b_g^2}\leq0.
\end{eqnarray}
Using the same arguments in the proof of Theorem 6, we can obtain $\lim_{L\rightarrow +\infty}\|e_{L}^{*}\|=0$, that completes the proof of Theorem 7.

Evaluation of the output weights in Theorem 7 is straightforward with the use of Moore-Penrose generalized inverse \cite{Lancaster1985}. Unfortunately, this method is infeasible for large-scale data analytics. To solve this problem, we suggest a trade-off solution with window shifting concept in the global least squares method (i.e., we only optimize a part of the output weights after the number of the hidden nodes exceeds a given window size). Indeed, this selective  scheme for evaluating the output weights is meaningful and significant for dealing with large-scale data processing. Similar to the proof of Theorem 7, the resulting SCN shares the universal approximation property. Here, we state this result in the following Theorem 8 and omit the  detailed proof. 

\textbf{Theorem 8.} Suppose that  span($\Gamma$) is dense in $L_2$ space and $\forall g\in \Gamma$, $0<\|g\|<b_g$ for some $b_g\in \mathds{R}^{+}$. Given $0<r<1$ and a nonnegative real number sequence $\{\mu_L\}$ with $\lim_{L\rightarrow+\infty}\mu_L=0$ and $\mu_L\leq (1-r)$. For a given window size $K$ and  $L=1,2,\ldots$, denoted by
\begin{equation}
\delta_{L}^{*}=\sum_{q=1}^{m}\delta_{L,q}^{*}, \delta_{L,q}^{*}=(1-r-\mu_L)\|e_{L-1,q}^{*}\|^2, q=1,2,...,m.
\end{equation}
If the random basis function $g_L$ is generated to satisfy the following inequalities:
\begin{equation}\label{step5}
%\sum_{q=1}^{m}\langle e_{L-1,q}^{*},g_L\rangle^2\geq b_g^2\delta_L^{*},
\langle e_{L-1,q}^{*},g_L\rangle^2\geq b_g^2\delta_{L,q}^{*}, q=1,2,...,m,
\end{equation}
and, as $L\leq K$, the output weights are evaluated by 
\begin{equation}\label{step6}
[\beta_1^{*}, \beta_2^{*},\ldots,\beta_{L}^{*}]=\arg \min_{\beta}\|f-\sum_{j=1}^{L}\beta_jg_j\|;
\end{equation}
Otherwise (i.e.,$L>K$), the output weights will be selectively evaluated (i.e., keep $\beta_1^{*},\ldots,\beta_{L-K}^{*}$ unchanged and renew the left $\beta_{L-K+1}, \ldots,\beta_{L}$) by 
\begin{equation}\label{step7}
[\beta_{L-K+1}^{*},\beta_{L-K+2}^{*},\ldots,\beta_{L}^{*}]=\arg \min_{\beta_{L-K+1}, \ldots,\beta_{L}}\|f-\sum_{j=1}^{L-K}\beta_j^{*}g_j-\sum_{j=L-K+1}^{L}\beta_jg_j\|.
\end{equation}
Then, we have $\lim_{L\rightarrow +\infty}\|f-f_L^{*}\|=0,$ where $f_L^{*}=\sum_{j=1}^{L}\beta_{j}^{*}g_j$, $\beta_{j}^{*}=[\beta^{*}_{j,1},\ldots,\beta^{*}_{j,m}]^{\mathrm{T}}$.

\subsection{Algorithm Description}
This subsection details the  proposed SC algorithms, i.e., SC-I, SC-II and SC-III, which are associated with Theorems 6, 8 and 7, respectively. In general, the main components of our proposed SC algorithms can be summarized as follows:
\begin{itemize}
  \item \textbf{Configuration of Hidden Parameters }: Randomly assigning the input weights and biases to meet the constraint (\ref{step1}) ((\ref{step3}) or (\ref{step5})), then generating a new hidden node and adding it to the current learner model.
  \item \textbf{Evaluation of Output Weights}: Constructively or selectively determining the output weights of the current learner model. 
\end{itemize}
\begin{table}[h!]\label{sc1}
\begin{center}
\begin{tabular}{lll}
\toprule
\textbf{Algorithm SC-I} \\
\midrule
Given inputs $X=\{x_1,x_2,\ldots,x_N\}$, $x_i\in \mathds{R}^{d}$ and outputs $T=\{t_1,t_2,\ldots,t_N\}$, $t_i\in \mathds{R}^{m}$; \\
Set maximum number of hidden nodes $L_{max}$, expected error tolerance $\epsilon$, maximum times\\ of random configuration $T_{max}$; Choose a set of positive scalars $\Upsilon\!=\{\lambda_{min}\!:\!\Delta\lambda\!:\!\lambda_{max}\}$;\\
\midrule
\textbf{1.} Initialize $e_0:=[t_1,t_2,\ldots,t_N]^\mathrm{T}$, $0<r<1$, two empty sets $\Omega$ and $W$;\\
\textbf{2.} \textbf{While} $L\leq L_{max}$ AND $\|e_0\|_{F}>\epsilon$, \textbf{Do}\\
\hspace{7.8mm}{\textbf{Phase 1: Hidden Parameters Configuration (Step 3-17)}} \\
\textbf{3.}\:\:\:\:\:\:\:\textbf{For} $\lambda \in \Upsilon$, \textbf{Do}\\
\textbf{4.}\:\:\:\:\:\:\:\:\:\:\:\:\:\:\textbf{For} $k=1,2\ldots,T_{max}$, \textbf{Do}\\
\textbf{5.}\:\:\:\:\:\:\:\:\:\:\:\:\:\:\:\:\:\:\:\:\:Randomly assign $\omega_L$ and $b_L$ from $[-\lambda,\lambda]^d$ and $[-\lambda,\lambda]$, respectively;\\
\textbf{6.}\:\:\:\:\:\:\:\:\:\:\:\:\:\:\:\:\:\:\:\:\:Calculate $h_{L}$, $\xi_{L,q}$ based on Eq. (\ref{hiddennode}) and (\ref{factor1}), and $\mu_L=(1-r)/(L+1)$;\\
\textbf{7.}\:\:\:\:\:\:\:\:\:\:\:\:\:\:\:\:\:\:\:\:\:\textbf{If}\:\:\:$min\{\xi_{L,1},\xi_{L,2},...,\xi_{L,m}\}\geq 0$\\
\textbf{8.}\:\:\:\:\:\:\:\:\:\:\:\:\:\:\:\:\:\:\:\:\:\:\:\:\:\:\:\textbf{Save} $w_L$ and $b_L$ in $W$, $\xi_L=\sum_{q=1}^{m}\xi_{L,q}$ in $\Omega$, respectively;\\
\textbf{9.}\:\:\:\:\:\:\:\:\:\:\:\:\:\:\:\:\:\:\:\:\:\textbf{Else} go back to \textbf{Step 4}\\
\textbf{10.}\:\:\:\:\:\:\:\:\:\:\:\:\:\:\:\:\:\:\:\textbf{End If}\\
\textbf{11.}\:\:\:\:\:\:\:\:\:\:\:\:\:\textbf{End For}\:(corresponds to \textbf{Step 4}) \\
\textbf{12.}\:\:\:\:\:\:\:\:\:\:\:\:\:\textbf{If}\:\:$W$ is not empty\\
\textbf{13.}\:\:\:\:\:\:\:\:\:\:\:\:\:\:\:\:\:\:\:Find $w_L^{*}$, $b_L^{*}$  that maximize $\xi_L$ in $\Omega$, and set $H_L=[h^*_1,h^*_2,\ldots,h^*_L]$;\\
\textbf{14.}\:\:\:\:\:\:\:\:\:\:\:\:\:\:\:\:\:\:\:\textbf{Break} (go to \textbf{Step 18}); \\
\textbf{15.}\:\:\:\:\:\:\:\:\:\:\:\:\:\textbf{Else} randomly take $\tau\in (0,1-r)$, renew $r:=r+\tau$, return to \textbf{Step 4};\\
\textbf{16.}\:\:\:\:\:\:\:\:\:\:\:\:\:\textbf{End If}\\
\textbf{17.}\:\:\:\:\:\:\textbf{End For} (corresponds to \textbf{Step 3})\\
\hspace{8.9mm}{\textbf{Phase 2: Output Weights Determination}} \\
\textbf{18.}\:\:\:\:\:\:Calculate $\beta_{L,q}=(e_{L-1,q}^{\mathrm{T}}\cdot h^{*}_L)/(h_L^{*\mathrm{T}}\cdot h^{*}_L)$, \:$q=1,2,\ldots,m$;\\ \textbf{19.}\:\:\:\:\:\:\:\:\:\:\:\:\:\:\:\:\:\:\:\:\:\:\:\:$\beta_{L}=[\beta_{L,1},\ldots,\beta_{L,m}]^{\mathrm{T}}$;\\
\textbf{20.}\:\:\:\:\:\:\:\:\:\:\:\:\:\:\:\:\:\:\:\:\:\:\:\:$e_L=e_{L-1}-\beta_Lh_{L}^{*}$; \\
\textbf{21.}\:\:\:\:\:\:Renew $e_0:=e_L$; $L:=L+1$;\\
\textbf{22.} \textbf{End While}\\
\textbf{23.} \textbf{Return} $\beta_1, \beta_2, \ldots, \beta_L$, $\omega^{*}=[\omega^{*}_1,\ldots,\omega^{*}_L]$ and $b^{*}=[b^{*}_1,\ldots,b^{*}_L]$.\\
\bottomrule
\end{tabular}
\end{center}
\end{table}

Given a training dataset with inputs $X=\{x_1,x_2,\ldots,x_N\}$, $x_i=[x_{i,1},\ldots,x_{i,d}]^\mathrm{T}\in \mathds{R}^{d}$ and its corresponding outputs $T=\{t_1,t_2,\ldots,t_N\}$, where $t_i=[t_{i,1},\ldots,t_{i,m}]^\mathrm{T}\in \mathds{R}^{m}$, $i=1,\ldots,N$. Denoted by $e_{L-1}(X)=[e_{L-1,1}(X),e_{L-1,2}(X),\ldots,e_{L-1,m}(X)]^\mathrm{T}\in \mathds{R}^{N\times m}$ as the corresponding residual error vector before adding the $L$-th new hidden node, where $e_{L-1,q}(X)=[e_{L-1,q}(x_1),\ldots,e_{L-1,q}(x_N)]\in \mathds{R}^N$ with $q=1,2,\ldots,m$. Let
\begin{equation}\label{hiddennode}
h_L(X)=[g_{L}(w_L^\mathrm{T}x_1+b_L),g_{L}(w_L^\mathrm{T}x_2+b_L),\ldots,g_{L}(w_L^\mathrm{T}x_N+b_L)]^\mathrm{T},
\end{equation}
which stands for the activation of the $L$-th hidden node for the input $x_i$, $i=1,2,\ldots,N$. The (current) hidden layer output matrix can be expressed as  $H_L=[h_1,h_2,\ldots,h_L]$.

In practice, the target function is presented as a collection of input-output data pairs. So $\beta_{L,q}=\langle e_{L-1,q},g_L\rangle/\|g_L\|^2$ ($q=1,\ldots,m$) becomes
\begin{equation}
\beta_{L,q}=\frac{e_{L-1,q}(X)^{\mathrm{T}}\cdot h_{L}(X)}{h_{L}(X)^{\mathrm{T}}\cdot h_{L}(X)}, \:\:\:q=1,2,\ldots,m.
\end{equation}
For the sake of brevity,  we introduce a set of variables $\xi_{L,q}, q=1,2,...,m$ and use them in algorithm descriptions (pseudo codes):
\begin{equation}\label{factor1}
\xi_{L,q}=\left(\frac{\Big(e_{L-1,q}(X)^{\mathrm{T}}\cdot h_L(X)\Big)^2}{h_L(X)^{\mathrm{T}}\cdot h_L(X)}-(1-r-\mu_L)e_{L-1,q}(X)^{\mathrm{T}}e_{L-1,q}(X)\right).
\end{equation}
Recall that for a given window size $K$ in SC-II, as $L\leq K$,  the suboptimal solution $\beta^{*}=[\beta^{*}_1,\beta^{*}_2,\ldots,\beta^{*}_L]^{\mathrm{T}}\in \mathds{R}^{L\times m}$ can be computed using the standard least squares method, that is,
\begin{equation}
  \beta^{*}=\arg\min_{\beta}\|H_L\beta-T\|_{F}^2=H^{\dagger}_LT,
\end{equation}
where $H^{\dagger}_L$ is the Moore-Penrose generalized inverse \cite{Lancaster1985} and $\|\cdot\|_{F}$ represents the Frobenius norm.

As $L>K$, a portion of the output weights can be evaluated by 
\begin{equation}
\beta^{window}=\arg\min_{\beta}\|H_K\beta-T\|_{F}^2=H^{\dagger}_KT,
\end{equation}
where $\beta^{window}$ consists of the most recent $\beta_{L-K+1},\ldots,\beta_L$, $H_K$ is composed of  the last $K$ columns of $H$, i.e., $H_K=[h_{L-K+1},\ldots,h_L]$, and the left (previous) $\beta_1,\ldots,\beta_{L-K}$ remain unchanged. It is easy to see that SC-II and SC-III become consistent once $K\geq L_{max}$.

\begin{table}[h!]\label{sc2}
\begin{center}
\begin{tabular}{lcl}
\toprule
\textbf{Algorithm SC-II} \\
\midrule
Given the same items in \textbf{Algorithm SC-I}. Set window size $K<L_{max}$. \\
\midrule
\textbf{1.} Initialize $e_0:=[t_1,t_2,\ldots,t_N]^\mathrm{T}$, $0<r<1$, two empty sets $\Omega$ and $W$;\\
\textbf{2.} \textbf{While} $L\leq L_{max}$ AND $\|e_0\|_{F}>\epsilon$, \textbf{Do}\\
\textbf{3.}\:\:\:\:\:\:\:\:Proceed \textbf{Phase 1 of Algorithm SC-I}; \\
\textbf{4.}\:\:\:\:\:\:\:\:Obtain $H_L=[h^*_1,h^*_2,\ldots,h^*_L]$;\\
\textbf{5.}\:\:\:\:\:\:\:\:\textbf{If}\:\:$L\leq K$\\
\textbf{6.}\:\:\:\:\:\:\:\:\:\:\:\:\:\:\:\:Calculate $\beta^{*}=[\beta^{*}_1,\beta^{*}_2,\ldots,\beta^{*}_L]^{\mathrm{T}}=H^{\dagger}_LT$;\\
\textbf{7.}\:\:\:\:\:\:\:\:\textbf{Else} \\
\textbf{8.}\:\:\:\:\:\:\:\:\:\:\:\:\:\:\:\:Set $\tilde{H}_K=H_L(:,1:L-K)$, $H_K=H_L(:,L-K+1:L)$;\\
\textbf{9.}\:\:\:\:\:\:\:\:\:\:\:\:\:\:\:\:Retrieve $\beta^{previous}=[\beta^{*}_1,\beta^{*}_2,\ldots,\beta^{*}_{L-K}]^{\mathrm{T}}$;\\
\textbf{10.}\:\:\:\:\:\:\:\:\:\:\:\:\:\:Calculate $\beta^{window}=H^{\dagger}_K(T-\tilde{H}_K\beta^{previous})$;\\
\textbf{11.}\:\:\:\:\:\:\:\:\:\:\:\:\:\:Let $\beta^{*}=[\beta^{*}_1,\beta^{*}_2,\ldots,\beta^{*}_L]^{\mathrm{T}}:=\left[\!\!\!
                                         \begin{array}{c}
                                           \beta^{previous} \\
                                           \beta^{window} \\
                                         \end{array}
                                       \!\!\!\!\right]$;\\
\textbf{12.}\:\:\:\:\:\:\:\textbf{End If}\\
\textbf{13.}\:\:\:\:\:\:\:Calculate $e_L=e_{L-1}-\beta_L^{*}h^{*}_L$;\\
\textbf{14.}\:\:\:\:\:\:\:Renew $e_0:=e_L$; $L:=L+1$;\\
\textbf{15.} \textbf{End While}\\
\textbf{16.} \textbf{Return} $\beta^{*}_1, \beta^{*}_2, \ldots, \beta^{*}_L$, $\omega^{*}=[\omega^{*}_1,\ldots,\omega^{*}_L]$ and $b^{*}=[b^{*}_1,\ldots,b^{*}_L]$.\\
\bottomrule
\end{tabular}
\end{center}
\end{table}
\begin{table}[htbp]\label{sc-3}
\begin{center}
\begin{tabular}{lcl}
\toprule
\textbf{Algorithm SC-III} \\
\midrule
Given the same items in \textbf{Algorithm SC-I}.\\
\midrule
\textbf{1.} Initialize $e_0:=[t_1,t_2,\ldots,t_N]^\mathrm{T}$, $0<r<1$, two empty sets $\Omega$ and $W$;\\
\textbf{2.} \textbf{While} $L\leq L_{max}$ AND $\|e_0\|_{F}>\epsilon$, \textbf{Do}\\
\textbf{3.}\:\:\:\:\:\:\:\:Proceed \textbf{Phase 1 of Algorithm SC-I};\\
\textbf{4.}\:\:\:\:\:\:\:\:Obtain $H_L=[h^*_1,h^*_2,\ldots,h^*_L]$;\\
\textbf{5.}\:\:\:\:\:\:\:\:Calculate $\beta^{*}=[\beta^{*}_1,\beta^{*}_2,\ldots,\beta^{*}_L]^{\mathrm{T}}:=H^{\dagger}_LT$;\\
\textbf{6.}\:\:\:\:\:\:\:\:Calculate $e_L=e_{L-1}-\beta_L^{*}h^{*}_L$;\\
\textbf{7.}\:\:\:\:\:\:\:\:Renew $e_0:=e_L$; $L:=L+1$;\\
\textbf{8.} \textbf{End While}\\
\textbf{9.} \textbf{Return} $\beta^{*}_1, \beta^{*}_2, \ldots, \beta^{*}_L$, $\omega^{*}=[\omega^{*}_1,\ldots,\omega^{*}_L]$ and $b^{*}=[b^{*}_1,\ldots,b^{*}_L]$.\\
\bottomrule
\end{tabular}
\end{center}
\end{table}

\textbf{Remark 5.} In SC-I, we employ a scaling sigmoidal function as the hidden node activation function, of which the parameters $\Upsilon
:=\{\lambda_1:\Delta\lambda:\lambda_{max}\}$ play  important role for adaptively determining the scope setting of the random parameters $w$ and $b$. As indicated in Theorem 6, although the randomness is beneficial for fast configuration of the newly added hidden node, the searching area is potentially data dependent and should not be rigidly fixed. In fact, the $\lambda\in \Upsilon$ varies during the course of adding new hidden nodes, that will be demonstrated later in our simulations.

\textbf{Remark 6.} Except for the expected error tolerance $\epsilon$, termination of SC-I can be done by referring to $L_{max}$ in order to prevent the over-fitting phenomenon. It is worth mentioning that a larger value of $\sum_{q=1}^{m}\langle e_{L-1,q},g_L\rangle^2/\|g_L\|^2$ would probably lead to faster decreasing in the residual error. This is  the reason behind why we conduct $T_{max}$ times of random configuration for $w_L$ and $b_L$ and finally return a pair of $w^{*}$ and $b^{*}$ with the largest  $\xi_L$, as specified between \textbf{Step 3-17} in Algorithm SC-I. Obviously, such an algorithmic operation helps in building compact SCNs.

\textbf{Remark 7.} As the constructive process proceeds, difficulties involved in the random configuration for $w$ and $b$ become increasingly higher when the residual error becomes smaller. To overcome this, we take time-varying value of $r$ so that $\sum_{q=1}^{m}(1-r-\mu_L)e_{L-1,q}(X)^{\mathrm{T}}e_{L-1,q}(X)$ will  approach to zero and consequently create more opportunities to make $\xi_L\geq0$.

\section{Performance Evaluation}
This section reports some simulation results over four regression problems, including a function approximation and three real-world modelling tasks. Performance evaluation goes to learning, generalization and efficiency (computing time and the eventual number of hidden nodes required for achieving an expected training error tolerance). Comparisons among  our SC algorithms, Modified Quickprop (MQ) \cite{Kwok1997} and IRVFL \cite{LiandWang2016} are carried out. In this study, the sigmoidal activation function $g(x)=1/(1+\exp(-x))$ is used. % and all simulations were conducted in MATLAB 7.0 on a computer with 3.5-GB RAM and 2.4-GHz Intel Core 2 Duo processor. 
\subsection{Data Sets}
Four datasets are employed in our simulation studies, including a function approximation example, three real world regression datasets \textbf{stock}, \textbf{concrete} and \textbf{compactiv} downloaded from KEEL (Knowledge Extraction based on Evolutionary Learning) dataset repository \footnote{KEEL: http://www.keel.es/}.
\begin{itemize}
  \item DB 1 was generated by a real-valued function \cite{Tyukin2009}:
\begin{equation}
f(x) = 0.2e^{-(10x-4)^{2}}+0.5e^{-(80x-40)^{2}}+0.3e^{-(80x-20)^{2}}, \:x\in[0,1].
\end{equation}
The training dataset has 1000 points randomly generated from the uniform distribution [0,1]. The test set, of size 300, is generated from a regularly spaced grid on [0,1].
\item Data provided in \textbf{stock} are daily stock prices from January 1988 through October 1991, for ten aerospace companies. The task is to approximate the price of the 10th company by using the prices of the rest. The whole data set (DB2) contains 950 observations with nine input variables and one output variable. We randomly selected 75\% samples as the training set while the test set consists of the left 25\%.
    \item The third regression task (\textbf{concrete}) is to find the highly nonlinear functional relationship between concrete compressive strength and the ingredients (input features) including cement, blast furnace slag, fly ash, water, super plasticizer, coarse aggregate, fine aggregate, and age. The whole data set (DB 3) contains 1020 instances. Both the training and test sets are formed in the same manner as DB2.
  \item The fourth dataset (\textbf{compactiv}) comes from a real world application, that is, computer activity dataset describing the portion of time that CPUs run in user-mode, based on 8192 computer system activities collected from a Sun SPARCstation 20/712 with 2 CPUs and 128 MB of memory running in a multi-user university
department. Each system activity is evaluated using 21 system measures (as input features). Both the training and test data sets are formulated in the same way as done for DB2 and DB3.
\end{itemize}
Both the input and output attributes are normalized into [0,1], and all results reported in this paper take averages over 100 independent trials.  The maximum times of random configuration $T_{max}$ is set as 200. For the other parameters such as the maximum number of hidden nodes $L_{max}$, the expected error tolerance $\epsilon$ and the index $r$, we will specify their corresponding settings later.

Two metrics are used in our comparative studies, i.e., modelling accuracy and efficiency. The first one is the widely used Root Mean Squares Error:
\begin{equation}
RMSE=\left(\frac{1}{N}\sum_{i=1}^{N}[\sum_{j=1}^L\beta_jg(w_j^{\mathrm{T}}x_i+b_j)-t_i]^2\right)^{1/2},
\end{equation}
where $N$ is the number of samples (training or test) and $L$ represents the number of hidden nodes. 
%In passing, all comparisons among the examined methods were performed on the basis of the same network structure, aiming to show the advantages of our SC algorithms (on both learning and generalization capabilities) on constructing a random learner model of given topology.

For the sake of comparing the efficiency of different methods, we recorded both the time (measured in seconds) spent on building the learner model and the number of hidden nodes needed to achieve the expected error tolerance.
\subsection{Results}
Table 1 and Table 2 show the training and test results of MQ, IRVFL, SC-I, SC-II and SC-III on DB 1-4, in which average values and standard deviations are reported on the basis of 100 independent trials. In particular, we set the learning rate in MQ as 0.05 and its maximum iterative number as 200 for all the regression tasks. In IRVFL, the random parameters ($w$ and $b$) are taken from the uniform distribution over [-1,1]. Clearly, from Table 1,  that SC-I, SC-II and SC-III outperform other methods in terms of both learning and generalization. For each setting of $L$, the training errors of MQ and IRVFL are larger than SC algorithms, in which SC-III demonstrates the best performance compared against SC-I and SC-II for both  training and test datasets. In Table 2, both MQ and SC algorithms lead to sound performances in comparison to the results obtained from IRVFL. 

For efficiency comparisons,  it can be observed from Table 3 that MQ, IRVFL and SC-I all fail in achieving the expected training error tolerance ($\epsilon=0.05$) within an acceptable time period for DB1, whilst both IRVFL and SC-I could not successfully achieve  the given training error for DB2. Here, we only report the results for  DB1 and DB2  as the other two case studies share similar consequences. First, the missing values in Table 3 (marked as `-') for time cost and test performance are caused by the extremely slow learning rate for these  scenarios. With some practical consideration that if certain patience parameter mentioned in \cite{Kwok1997} is given, the learning phase of MQ, IRVFL and SC-I algorithm will be terminated to avoid meaningless time cost. Consequently, it seems impossible for those methods to meet the expected error tolerance in a reasonable time slot. The proposed SC-II and SC-III algorithms, however, perform quite well and achieve the specified training error bound with few hidden nodes. On the other hand, compared against MQ method, SC-II and SC-III algorithms generate less number of hidden nodes and demonstrate better generalization performance. 

\begin{table}[h!]
\centering
\footnotesize
{\caption{Performance Comparison among MQ, IRVFL, SC-I, SC-II and SC-III on DB 1. $K=15$ was used in SC-II. }\label{tab:1}}
\begin{center}
\begin{tabular}{c|cc|cc}\hline
\multirow{2}*{Algorithms} &\multicolumn{2}{c}{Training}\vline  & \multicolumn{2}{c}{Test}\\
\cline{2-5}
   &$L=25$ &  $L=50$ &  $L=25$& $L=50$  \\
\hline
MQ &  0.1031$\pm$0.0001   & 0.1030$\pm$0.0001 & 0.1011$\pm$0.0003 & 0.1011$\pm$0.0003   \\
IRVFL             &  0.1630$\pm$0.0008&  0.1626$\pm$0.0005& 0.1622$\pm$0.0012 & 0.1617$\pm$0.0008  \\
SC-I             &  0.0927$\pm$0.0020&  0.0887$\pm$0.0018& 0.0912$\pm$0.0021 & 0.0870$\pm$0.0019 \\
SC-II   & \textbf{0.0435}$\pm$\textbf{0.0061}  &  \textbf{0.0366}$\pm$\textbf{0.0049}&  \textbf{0.0411}$\pm$\textbf{0.0064}& \textbf{0.0337}$\pm$\textbf{0.0049}\\
SC-III           &  \textbf{0.0332}$\pm$\textbf{0.0065} &  \textbf{0.0097}$\pm$\textbf{0.0036}& \textbf{0.0308}$\pm$\textbf{0.0060} & \textbf{0.0100}$\pm$\textbf{0.0033}  \\
\hline
\end{tabular}
\end{center}
\end{table}
\begin{table}[h!]
\centering
\footnotesize
{\caption{Performance Comparison among MQ, IRVFL and SC-I, SC-II and SC-III on DB 2, DB3 and DB4. $K=15$ was used in SC-II. }\label{tab:2}}
\begin{center}
\begin{tabular}{c|c|cc|cc}\hline
\multirow{2}*{Data Sets} &\multirow{2}*{Algorithms} &\multicolumn{2}{c}{Training}\vline  & \multicolumn{2}{c}{Test}\\
\cline{3-6}
   &&$L=25$ &  $L=50$ &  $L=25$& $L=50$ \\
\hline
\multirow{5}*{DB2} & MQ &  0.0624$\pm$0.0058  & 0.0410$\pm$0.0014&  0.0611$\pm$0.0061 & 0.0407$\pm$0.0017     \\
                                              &IRVFL  &  0.2121$\pm$0.0260 &  0.1853$\pm$0.0248& 0.2057$\pm$0.0268 & 0.1787$\pm$0.0237 \\
                                              &SC-I             &   0.0963$\pm$0.0032& 0.0881$\pm$0.0026& 0.0925$\pm$0.0031 & 0.0851$\pm$0.0024\\
                                              &SC-II    & \textbf{0.0437}$\pm$\textbf{0.0014}  &  \textbf{0.0395}$\pm$\textbf{0.0008}&  \textbf{0.0427}$\pm$\textbf{0.0017} & \textbf{0.0391}$\pm$\textbf{0.0010}  \\
                                              &SC-III           &  \textbf{0.0409}$\pm$\textbf{0.0010} &  \textbf{0.0327}$\pm$\textbf{0.0007}& \textbf{0.0403}$\pm$\textbf{0.0012} & \textbf{0.0347}$\pm$\textbf{0.0012} \\
\hline
\multirow{5}*{DB3} & MQ &  0.1096$\pm$0.0042  & 0.0910$\pm$0.0014&  0.0999$\pm$0.0048 & 0.0869$\pm$0.0021     \\
                                              &IRVFL  &  0.2045$\pm$0.0189 &  0.1929$\pm$0.0135& 0.2109$\pm$0.0214 & 0.1983$\pm$0.0166 \\
                                              &SC-I             &  0.1401$\pm$ 0.0027&   0.1358$\pm$0.0022& 0.1315$\pm$0.0026& 0.1258$\pm$0.0017\\
                                              &SC-II    & \textbf{0.0994}$\pm$\textbf{0.0018}  &  \textbf{0.0943}$\pm$\textbf{0.0011}&  \textbf{0.0925}$\pm$\textbf{0.0029} & \textbf{0.0874}$\pm$\textbf{0.0018}  \\
                                              &SC-III           &  \textbf{0.0969}$\pm$\textbf{0.0013} &  \textbf{0.0835}$\pm$\textbf{0.0012}& \textbf{0.0898}$\pm$\textbf{0.0020} & \textbf{0.0850}$\pm$\textbf{0.0025} \\
\hline
\multirow{5}*{DB4} & MQ &  0.0840$\pm$0.0052  & 0.0600$\pm$0.0071&  0.0848$\pm$0.0053 & 0.0624$\pm$0.0075     \\
                                              &IRVFL  &  0.2002$\pm$0.0391 &  0.1924$\pm$0.0283& 0.1958$\pm$0.0386 & 0.1882$\pm$0.0281 \\
                                              &SC-I             &  0.1207$\pm$0.0036& 0.1137$\pm$0.0038& 0.1169$\pm$0.0038&0.1105$\pm$ 0.0040\\
                                              &SC-II    & \textbf{0.0760}$\pm$\textbf{0.0034}  &  \textbf{0.0579}$\pm$\textbf{0.0029}&  \textbf{0.0773}$\pm$\textbf{0.0039} & \textbf{0.0593}$\pm$\textbf{0.0036} \\
                                              &SC-III           &  \textbf{0.0678}$\pm$\textbf{0.0038} &  \textbf{0.0394}$\pm$\textbf{0.0016}& \textbf{0.0697}$\pm$\textbf{0.0044} & \textbf{0.0418}$\pm$\textbf{0.0021} \\
\hline
\end{tabular}
\end{center}
\end{table}

\begin{table}[h!]
\centering
\footnotesize
{\caption{Efficiency comparison among MQ, IRVFL and SC-I, SC-II and SC-III on DB 1 and DB2. Window size $K=15$ was used in SC-II.}\label{tab:3}}
\begin{center}
\begin{tabular}{c|c|cc|c}\hline
\multirow{2}*{Data Sets} &\multirow{2}*{Algorithms} &\multicolumn{2}{c}{Efficiency ($\epsilon=0.05$)}\vline  & \multirow{2}*{Test}\\
\cline{3-4}
   & &Time (\emph{s})& Nodes \\
\hline
\multirow{5}*{DB1} & MQ &  - & - &-  \\
                                              &IRVFL  & -&  - &-\\
                                              &SC-I             &  - & -&-\\
                                              &SC-II    & 0.4029$\pm$0.1690  &  26.6700$\pm$8.7514 &0.0463$\pm$0.0021\\
                                              &SC-III           &  \textbf{0.2737}$\pm$\textbf{0.0719} &  \textbf{19.9000}$\pm$\textbf{3.6167}&\textbf{0.0435}$\pm$\textbf{0.0047} \\
\hline
\multirow{5}*{DB2} & MQ & 1.3157$\pm$0.6429  &  34.1809$\pm$3.5680 &0.0532$\pm$0.0175\\
                                              &IRVFL  &  -&  - &-\\
                                              &SC-I             &  - & -&-\\
                                              &SC-II    & 0.2193$\pm$0.0454  &  16.7200$\pm$2.1513 &0.0488$\pm$0.0016\\
                                              &SC-III           &  \textbf{0.2083}$\pm$\textbf{0.0317} &  \textbf{16.3600}$\pm$\textbf{1.5987}&\textbf{0.0483}$\pm$\textbf{0.0020} \\
\hline
\end{tabular}
\end{center}
\end{table}

\begin{figure*}[h!]
\centering
\subfigure[]{\includegraphics[width=0.46\textwidth]{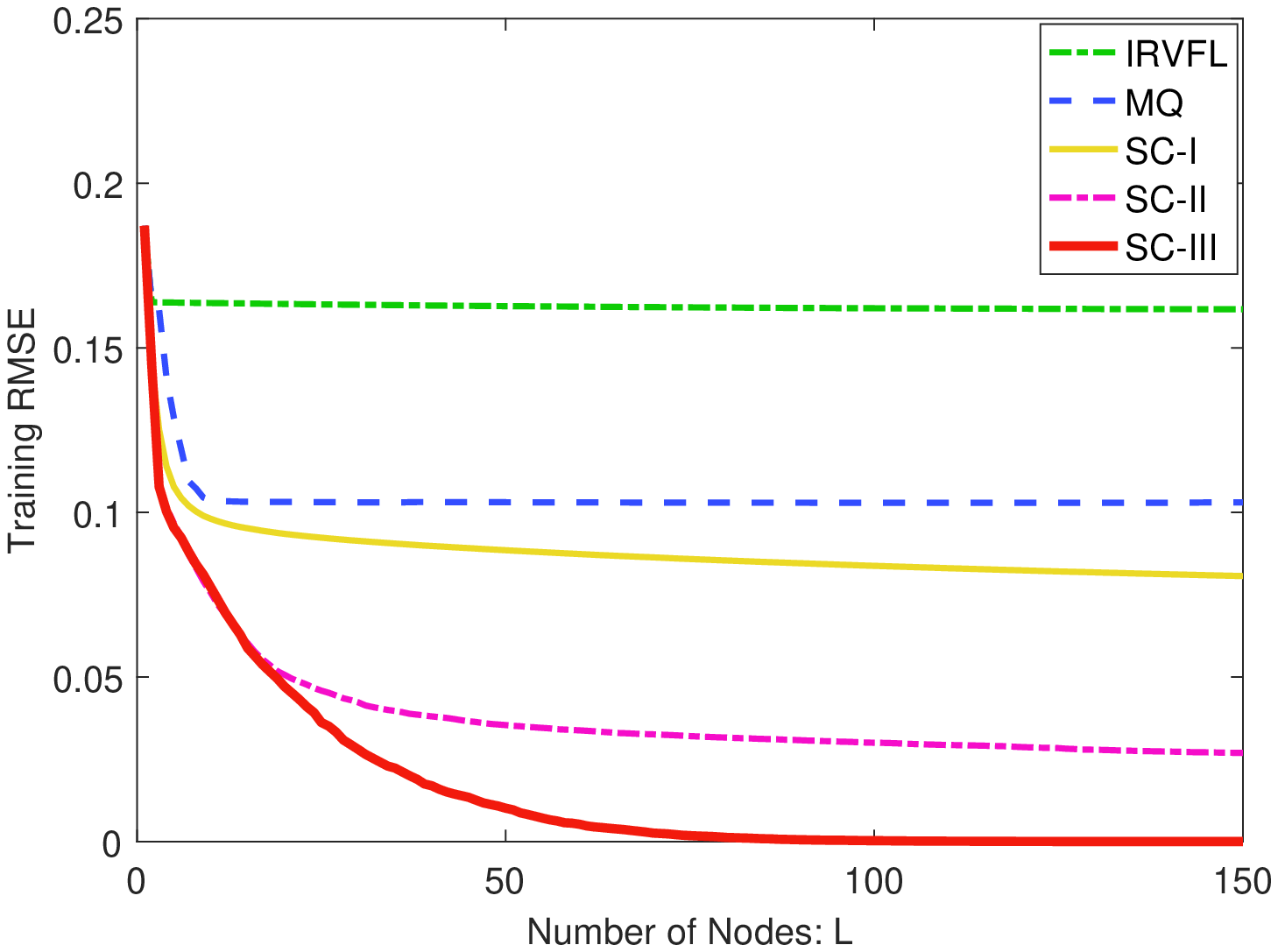}}
\subfigure[]{\includegraphics[width=0.46\textwidth]{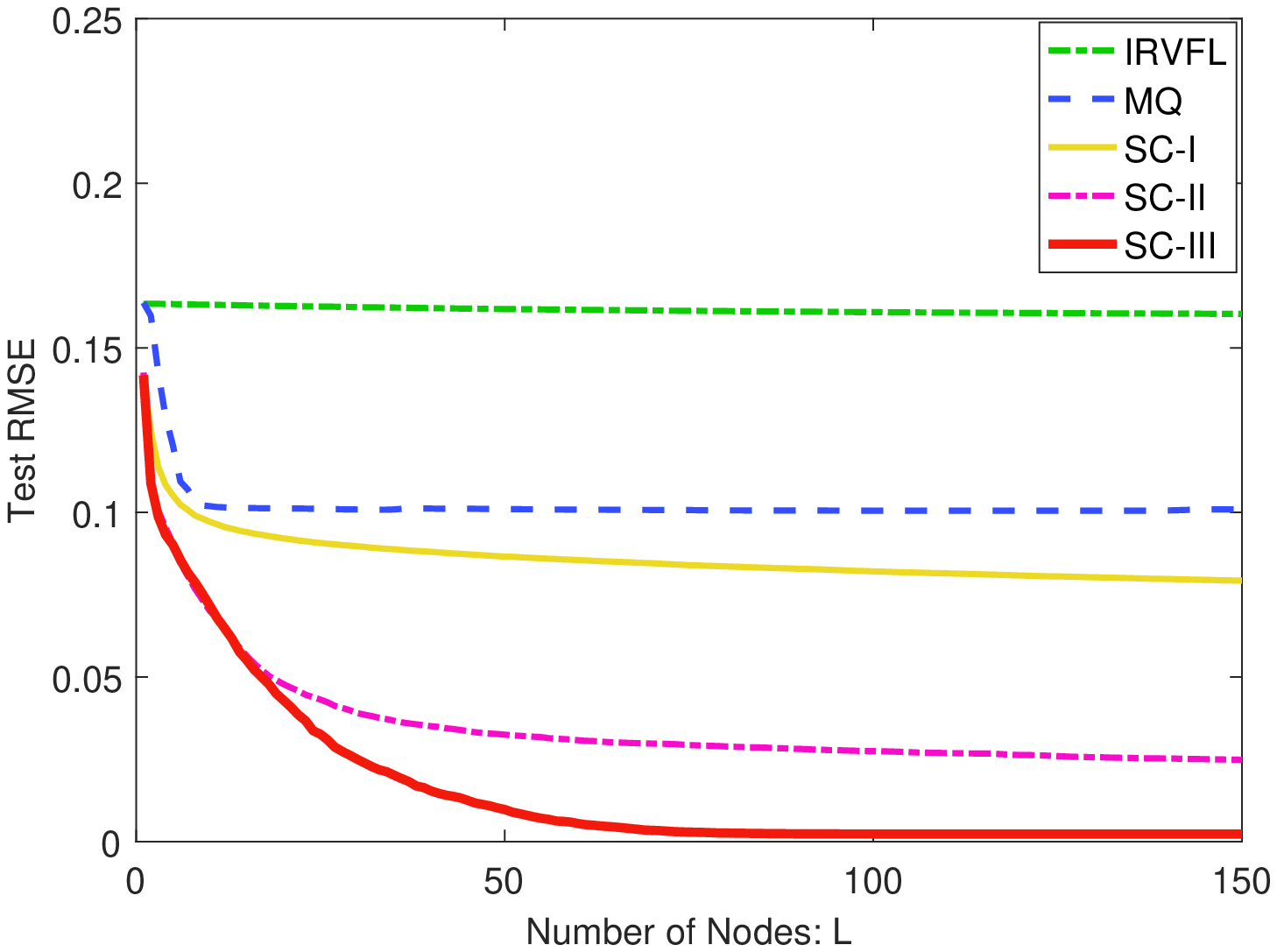}}
\caption{Performance of Modified QuickProp (MQ), IRVFL, SC-I, SC-II and SC-III with 150 additive nodes on DB 1: (a) Average training RMSE and (b) Average test RMSE}\label{fig:1}
\end{figure*}
\begin{figure*}[h!]
\centering
\subfigure[]{\includegraphics[width=0.46\textwidth]{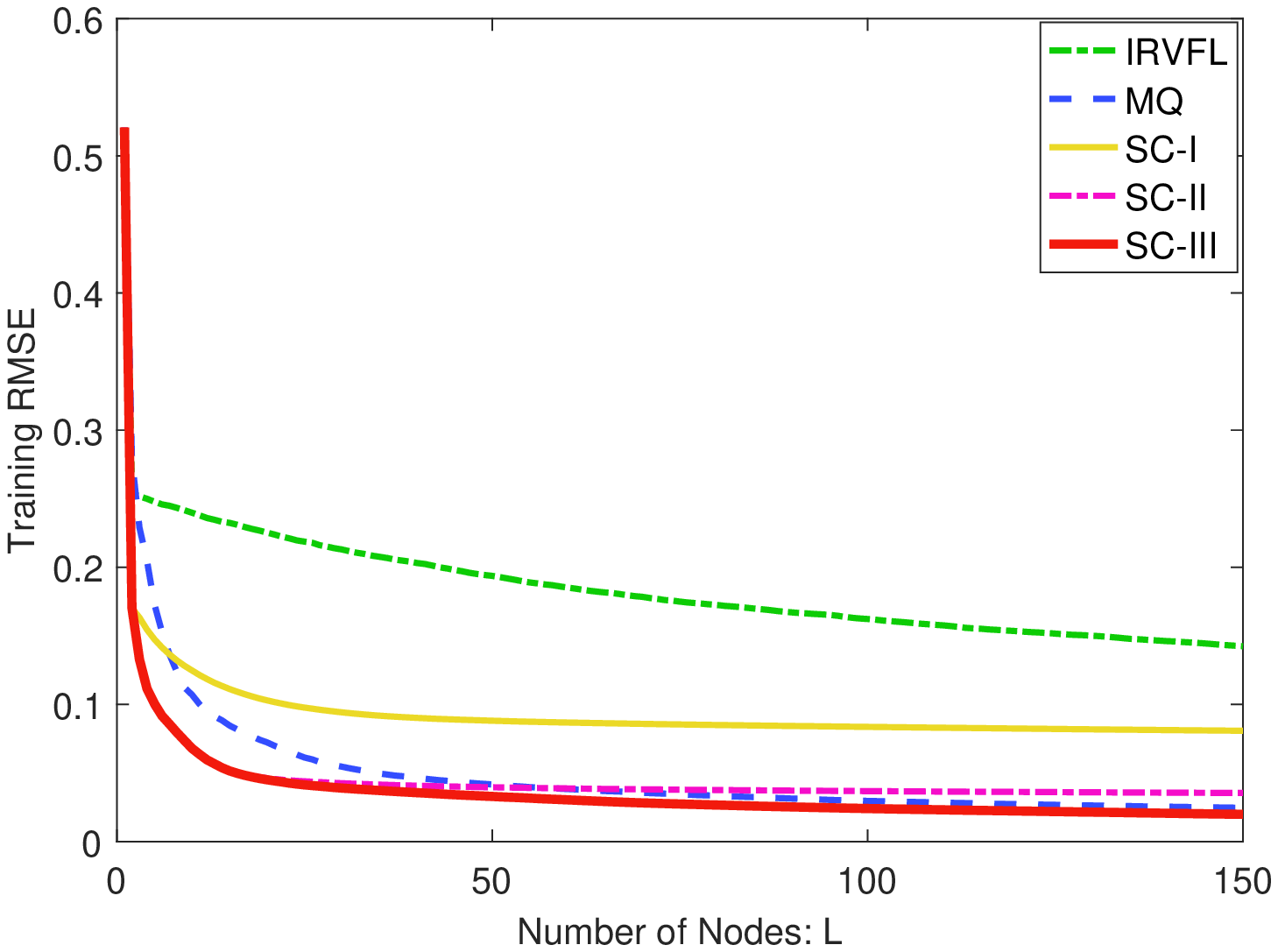}}
\subfigure[]{\includegraphics[width=0.46\textwidth]{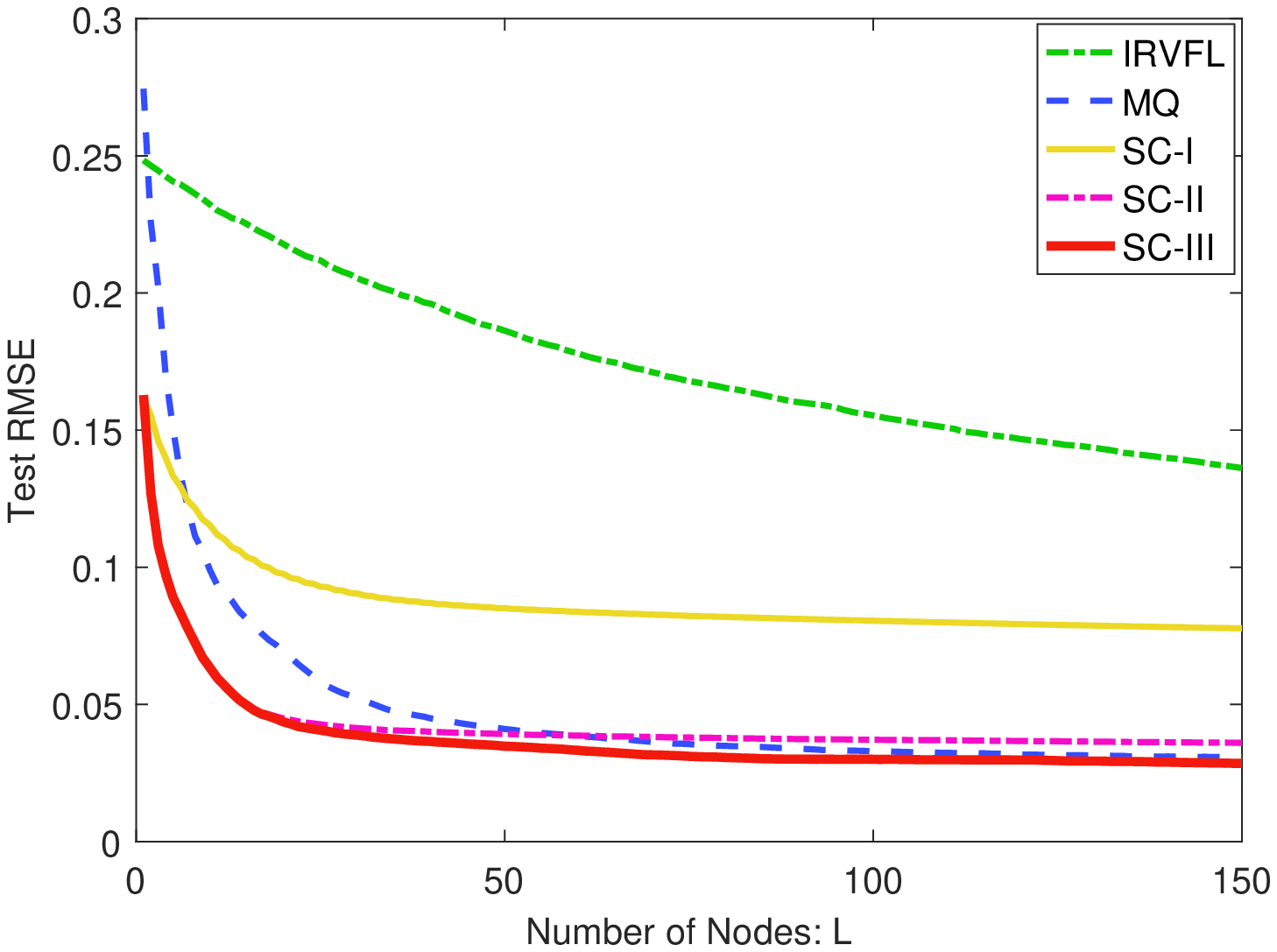}}
\caption{Performance of Modified QuickProp (MQ), IRVFL, SC-I, SC-II and SC-III with 150 additive nodes on DB 2: (a) Average training RMSE and (b) Average test RMSE}\label{fig:2}
\end{figure*}

The advantages of the proposed SC algorithms are clearly visible from Figure 1 and Figure 2, in which the curves are plotted based on the average values over 100 independent runs. The following observations can be made from these results.
\begin{itemize}
  \item In Figure 1, it can be observed that the error decreasing rates of our proposed SC algorithms are quite higher than the other two methods. At the same time, their corresponding test errors also keep decreasing when adding new hidden nodes. Furthermore, similar to the results reported in Table 1, SC-II and SC-III perform much better than SC-I, which has slower decreasing rate (rather than MQ and IRVFL). It should be noted that window size $K=15$ is used in SC-II, so that the error decreasing trend of SC-II and SC-III should be consistent with each other when $L\leq 15$. This fact has been clearly illustrated in Figure 1.
  \item For MQ and IRVFL algorithms on DB 1, both the training and test errors stabilize at a certain level but still unacceptable (the real training RMSE is larger than 0.1). This extremely slow decreasing rate in learning makes it impossible to build a learner model with time constraint, even when $L$ takes a larger value. In comparison, an improved decreasing rate of SC-I seems obviously evident compared against MQ after adding 150 hidden nodes.
  \item In Figure 2, MQ, SC-II, and SC-III converge faster than the other two algorithms, of which IRVFL exhibits the slowest decreasing rate in both training and test phases. Although MQ in this case outperforms SC-I, its error decreasing rate is slower than SC-II and SC-III as $L\leq 50$. Indeed, there is a negligible difference among the curves of SC-II, SC-III and MQ when $L>50$. As a whole, the error curves of SC-III always stay at the bottom, which correspond to the numerical results reported in Table 3.
\end{itemize}

\begin{table}[h!]
\centering
\footnotesize
{\caption{Performance of SC-II with different window sizes on DB 1. $L=35$}\label{tab:4}}
\begin{tabular}{c|cc|cc|cc}\hline
\multirow{2}*{Window Size} &\multicolumn{2}{c|}{Training}  & \multicolumn{2}{c|}{Test}& \multicolumn{2}{c}{Efficiency ($\epsilon=0.05$)}\\
\cline{2-7}
   &Mean &  STD &  Mean& STD &Time(\emph{s}) &Nodes   \\
\hline
K=5&  0.0682  &  0.0049& 0.0672 & 0.0049 & 1.36 &99.99\\
K=10&  0.0530  &  0.0061 & 0.0516 & 0.0064 & 0.72 & 50.85\\
K=15&  0.0439  &  0.0062 & 0.0416 & 0.0068 & 0.34 & 25.07\\
K=20&  0.0364  &  0.0069 & 0.0338 & 0.0064 & 0.28 & 21.21\\
K=25&  0.0325  &  0.0073 & 0.0303 & 0.0067 & 0.26 & 20.29\\
K=30&  0.0267 &  0.0056 & 0.0252 & 0.0049 & 0.25 & 19.62\\
K=35&  0.0248 &  0.0069 & 0.0233 & 0.0062 & 0.25 & 19.75\\
\hline
\end{tabular}
\end{table}

\begin{table}[h!]
\centering
\footnotesize
{\caption{Performance of SC-II with different window sizes $K$ on DB 2. $L=25$}\label{tab:5}}
\begin{tabular}{c|cc|cc|cc}\hline
\multirow{2}*{Window Size} &\multicolumn{2}{c|}{Training}  & \multicolumn{2}{c|}{Test}& \multicolumn{2}{c}{Efficiency ($\epsilon=0.05$)}\\
\cline{2-7}
   &Mean &  STD &  Mean& STD &Time(\emph{s}) &Nodes   \\
\hline
K=5&  0.0613  &  0.0039& 0.0601 & 0.0037 & 0.88 & 66.42\\
K=10&  0.0492  &  0.0024 & 0.0493 & 0.0027 & 0.31& 23.48\\
K=15&  0.0443  &  0.0014 & 0.0436 & 0.0020& 0.21 & 16.62\\
K=20&  0.0420 &  0.0012& 0.0412& 0.0015 & 0.20& 16.16\\
K=25&  0.0411  &  0.0010 & 0.0405 & 0.0013 & 0.20 & 16.35\\
\hline
\end{tabular}
\end{table}

Table 4 and 5 report some results on robustness of the modelling performance with respect to  the window size $K$ in SC-II. It should be noted that we specified the number of hidden nodes being added in order to perform the robustness analysis. Furthermore, for the purpose of examining the efficiency, we set the training error tolerance for each task (the same as in Table 3) and recorded the real time cost and number of hidden notes required for achieving that tolerance. All of the results reported in Table 4 and 5 are averaged over 100 independent runs. For DB 1, seven cases ($K=5, 10,15, 20,25, 30, 35$) are considered and the number of hidden nodes is set as 35. It can been clearly seen that both training and test performance become better along with  increasing the value of $K$. Importantly, small $K$ values may lead to inferior results, for example, the training and test RMSEs with  $K=5$, $K=10$ and $K=15$ are out of the tolerance level. On the other hand, the efficiency of SC-II will become higher as the value of $K$ increases. Apart from the scenarios of $K=5$, $K=10$ and $K=15$, the difference among the remaining cases can be ignored that the number of hidden nodes is about 20 while the whole time cost is around 0.26$s$ (see Table 4). For DB 2, five cases ($K=5, 10, 15, 20, 25$) are examined while the number of hidden nodes is set as 25. Similarly,  all of the results, except for $K=5$ and $K=10$, are comparable. In particular, for $K=15,20,25$, the time cost and number of hidden nodes required for achieving the error tolerance ($\epsilon=0.05$) are around 0.2$s$ and 16, respectively. 
\begin{figure*}[h!]
\centering
\subfigure[IRVFL, $\lambda=1$]{\includegraphics[width=0.32\textwidth]{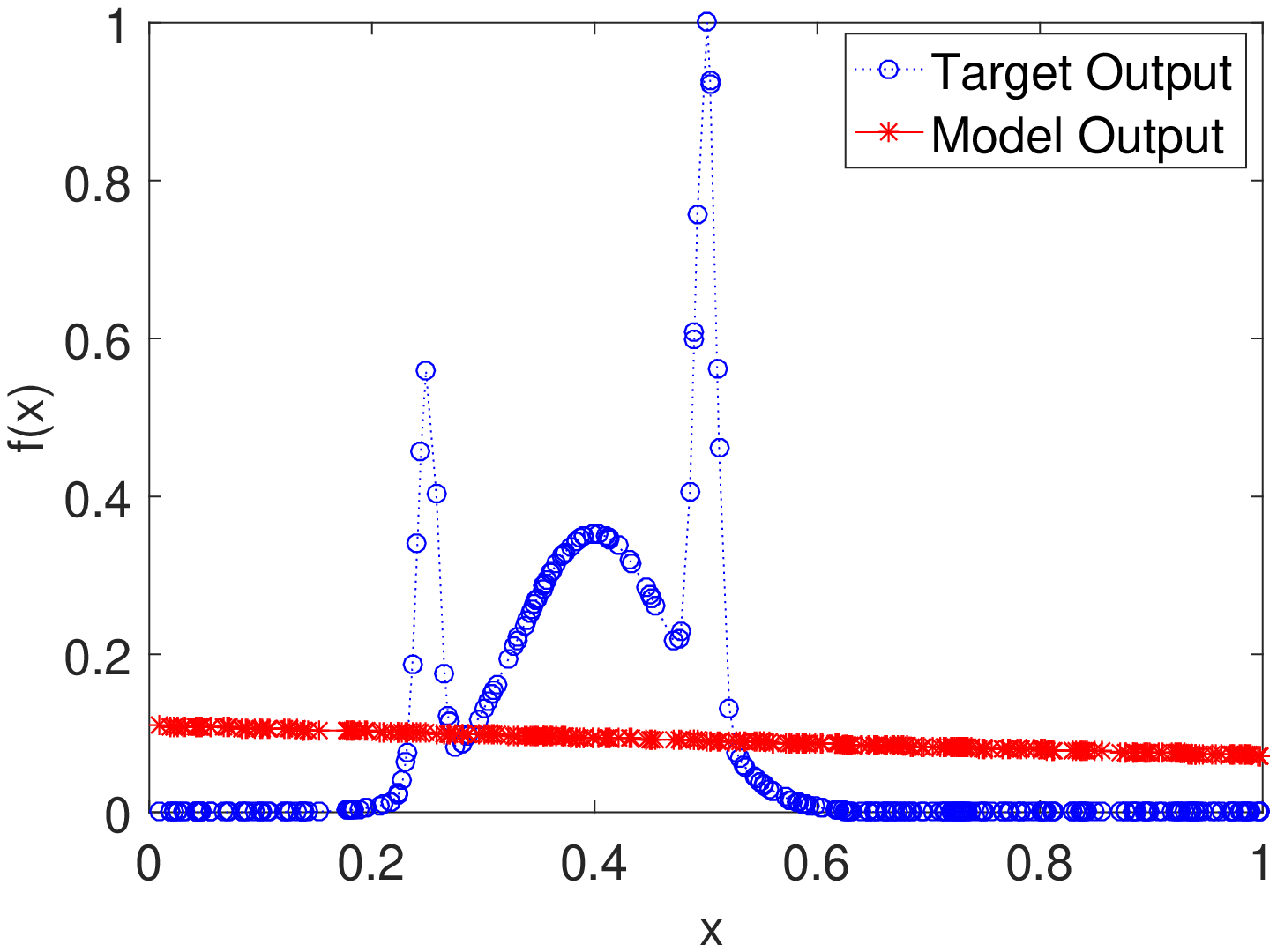}}
\subfigure[IRVFL, $\lambda=100$]{\includegraphics[width=0.32\textwidth]{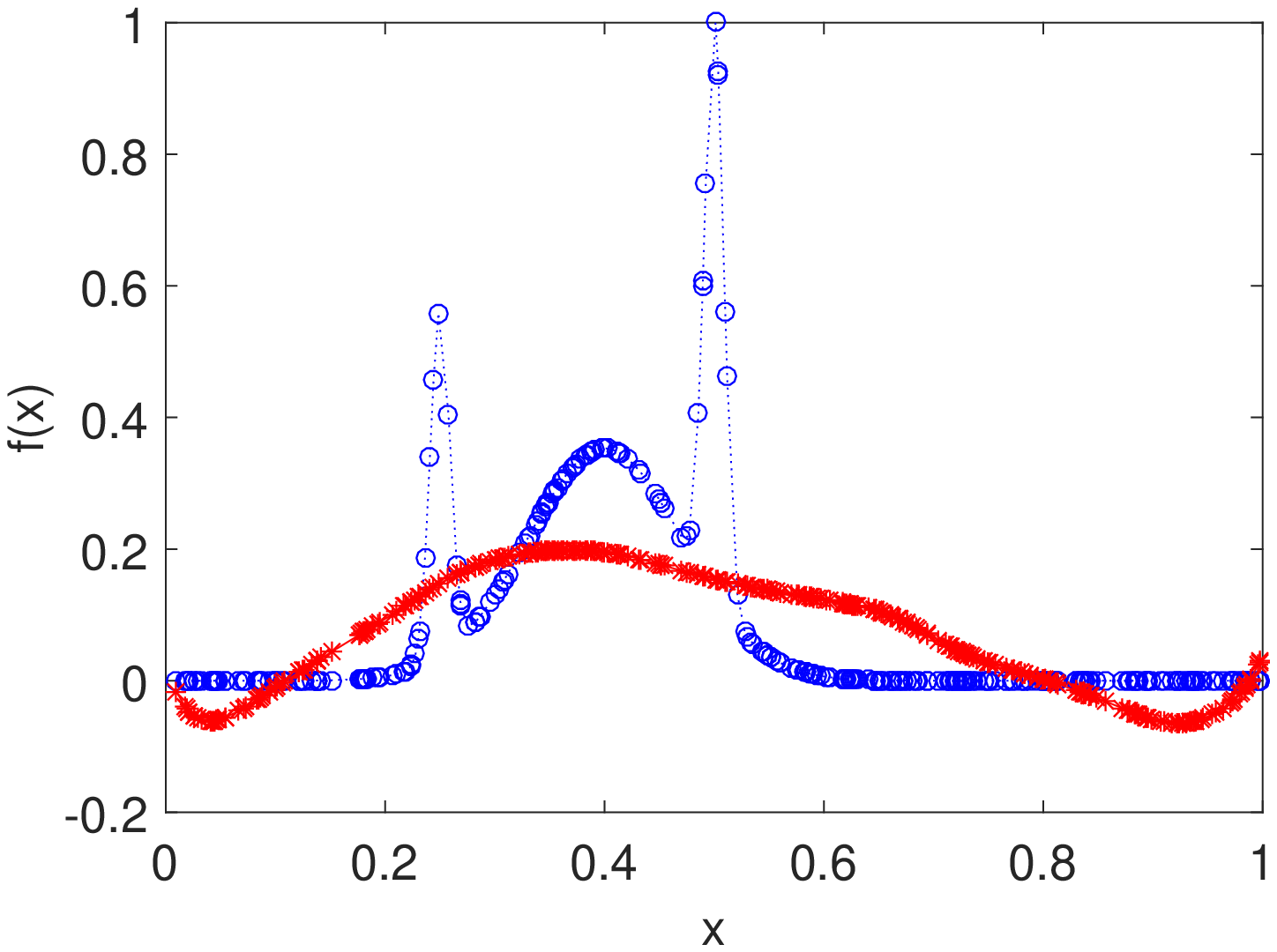}}
\subfigure[IRVFL, $\lambda=200$]{\includegraphics[width=0.32\textwidth]{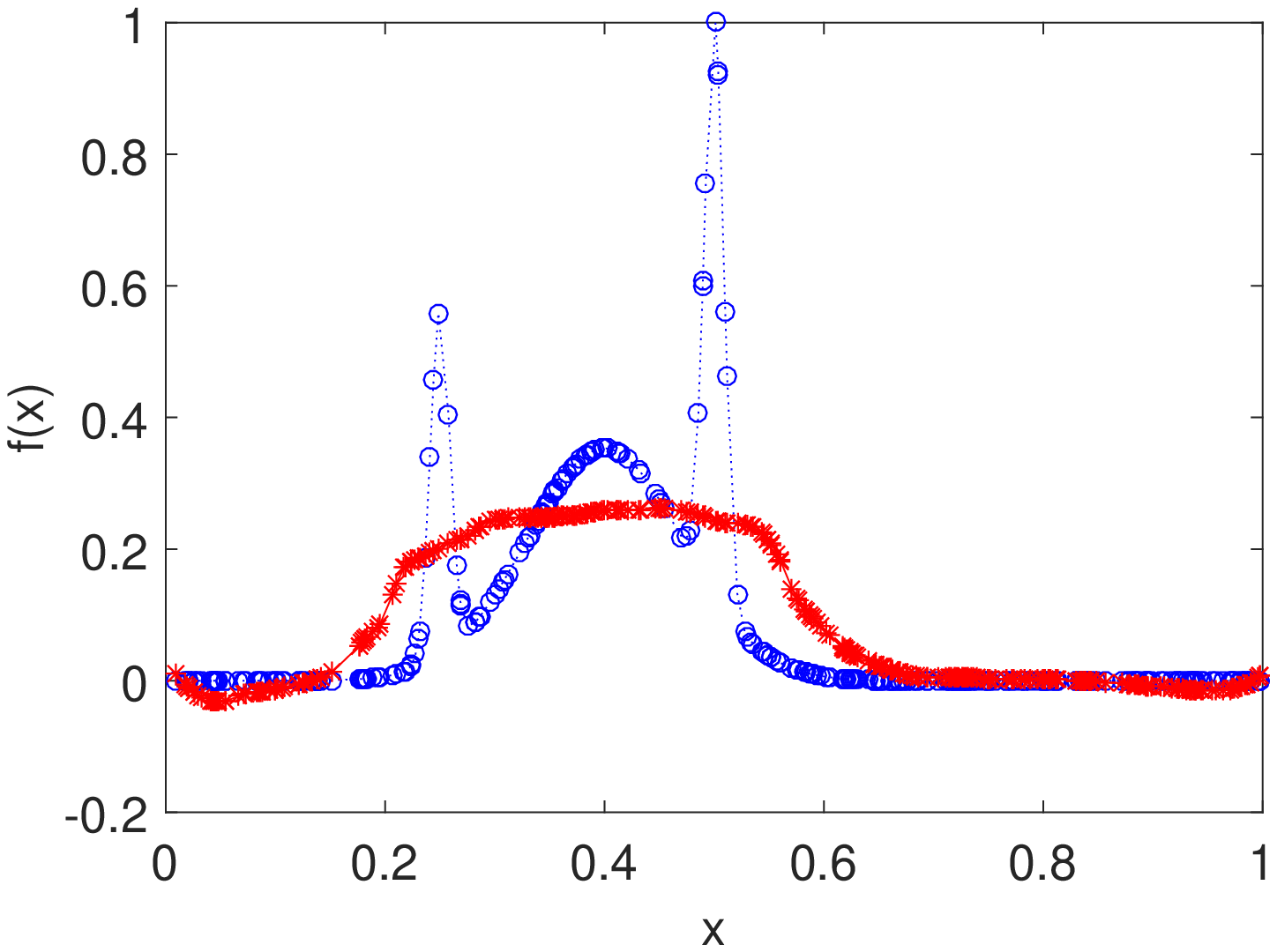}}
\subfigure[SC-I]{\includegraphics[width=0.32\textwidth]{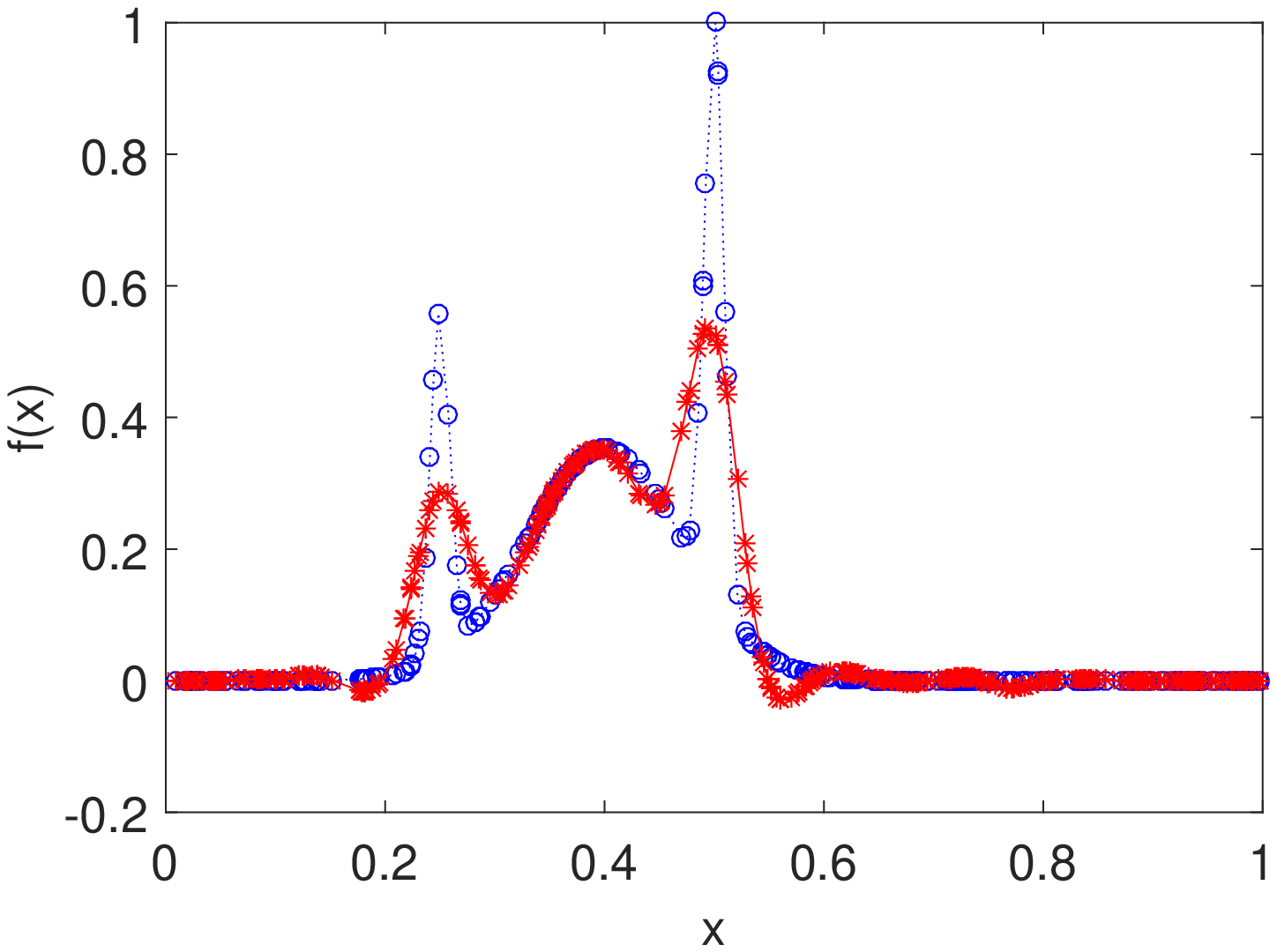}}
\subfigure[SC-II, $K=15$]{\includegraphics[width=0.32\textwidth]{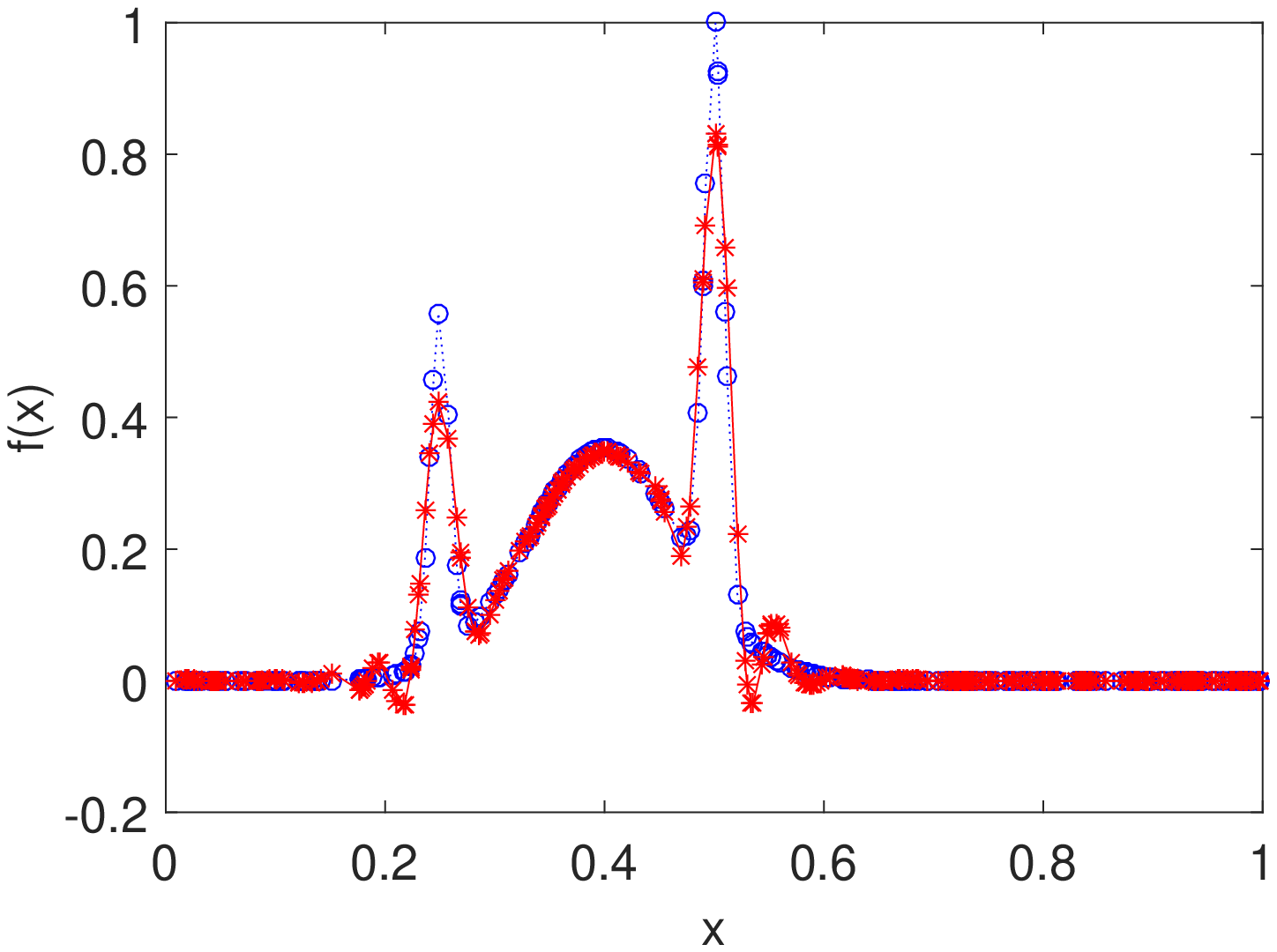}}
\subfigure[SC-III]{\includegraphics[width=0.32\textwidth]{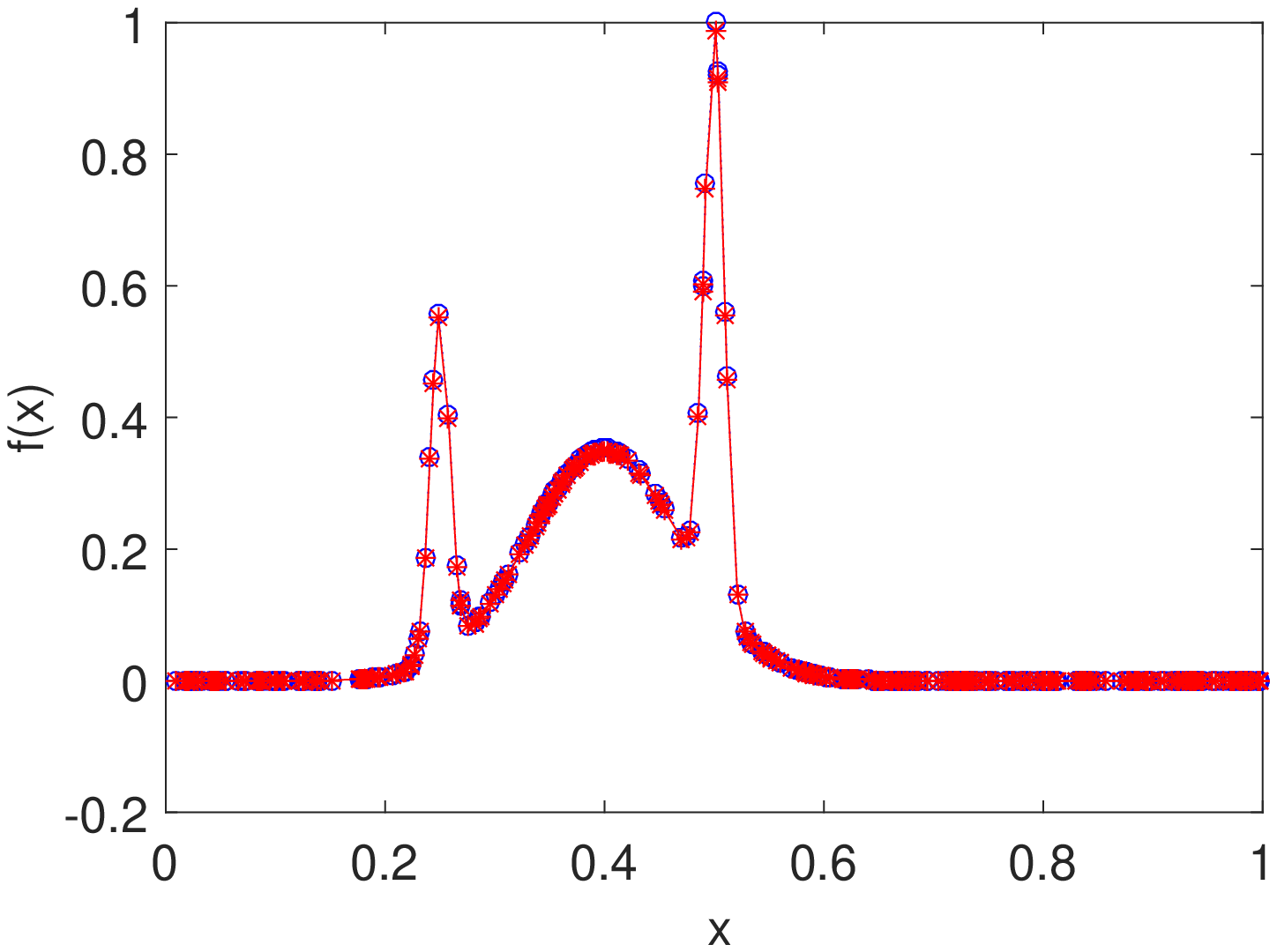}}
\caption{Approximation performance of IRVFL (with different setting of $\lambda$), SC I, SC-II and SC-III on DB 1}\label{fig:3}
\end{figure*}

Figure 3 depicts  the modelling performance (for test dataset) of IRVFL, SC-I, SC-II and SC-III on DB 1, respectively. Firstly, the importance of the scale factor $\lambda$ in the activation function, which directly determines the range of random parameters, is examined by performing different settings. Secondly, the infeasibility of IRVFL is illustrated in comparison with the proposed SC algorithms, where effectiveness and benefits from  the supervisory mechanism of SCNs can be clearly observed. The maximum number of the hidden nodes to be added is set as 100 for these four methods examined here. From Figure 3(a), (b), (c), it is apparent that IRVFL networks perform poorly. In fact, we attempted to add 20,000 hidden nodes to see the performance of IRVFL networks with different values of $\lambda$. Unfortunately, the final result is very similar to that depicted in Figure 3(c), which is  aligned with our findings in \cite{LiandWang2016}. In comparison, the test performance of SC-I shown in Figure 3(d) is much better than IRVFL, that corresponds to the records in Table 1 and the error changing curves in Figure 1, respectively. In addition, both SC-II and SC-III outperform SC-I as the recalculation of $\beta_L$ contributes a lot in constructing a universal approximator with a much more compact structure.
\begin{figure*}[h!]
\centering
\subfigure[Training]{\includegraphics[width=0.46\textwidth]{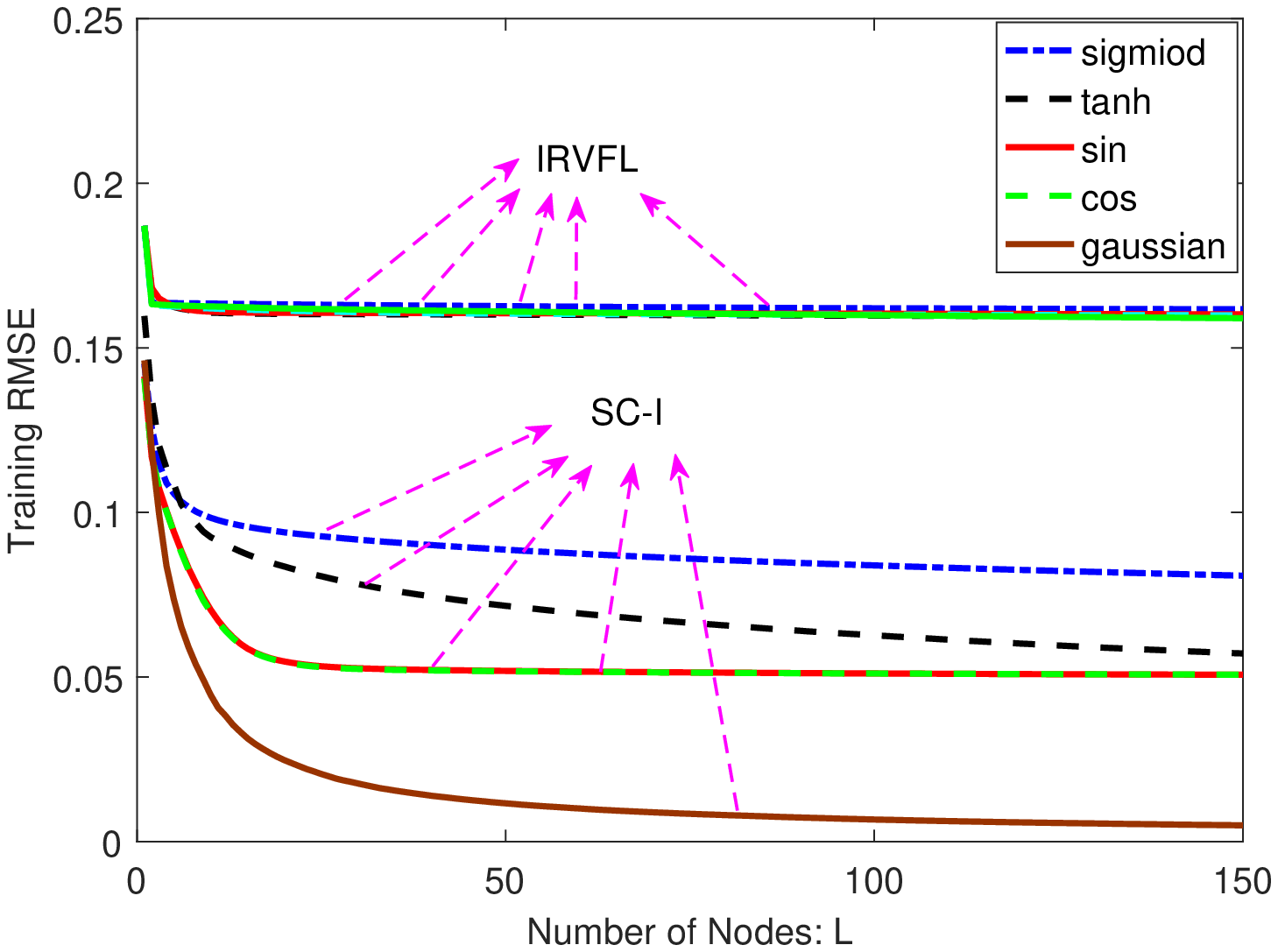}}
\subfigure[Test]{\includegraphics[width=0.46\textwidth]{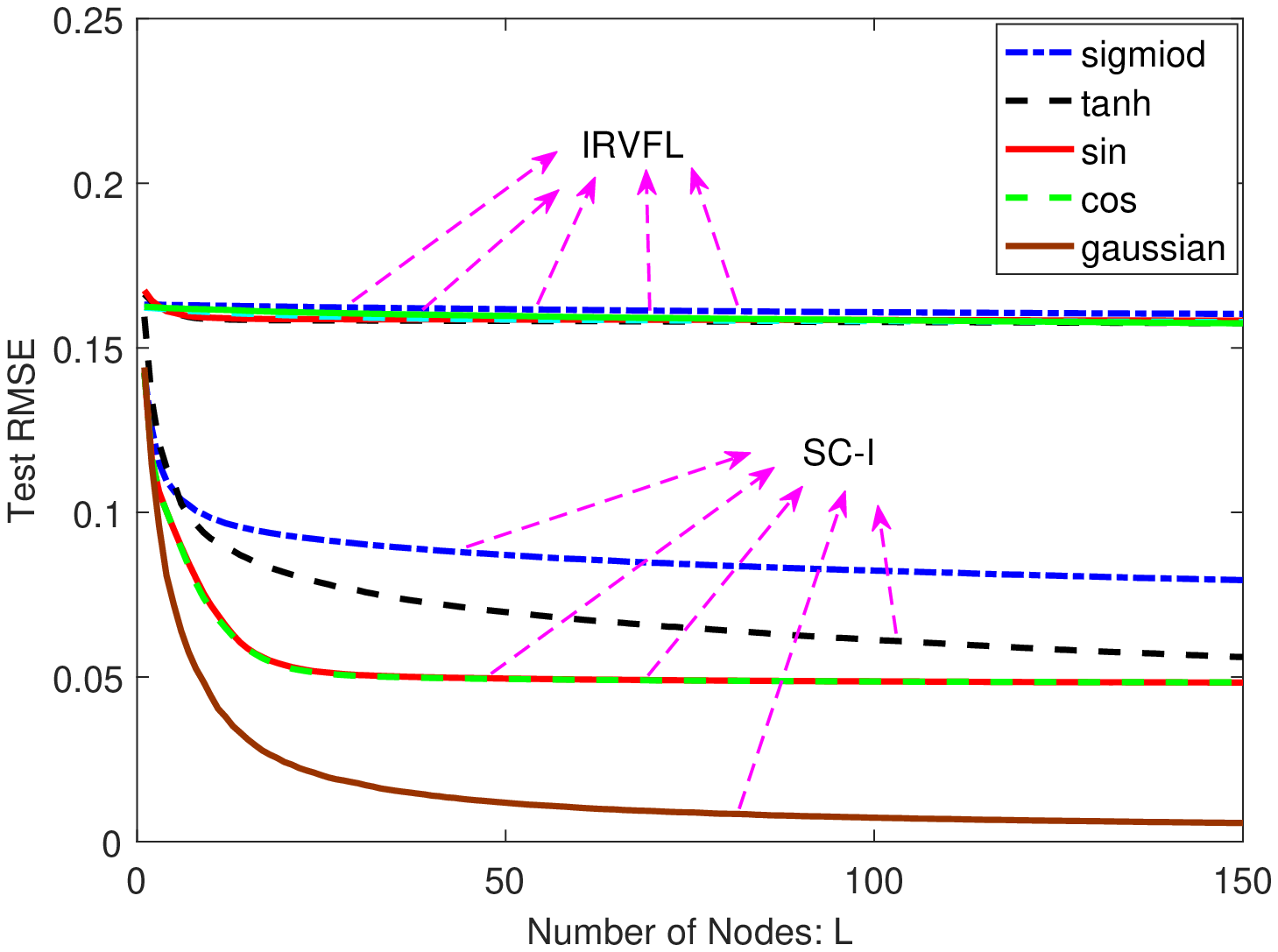}}
\caption{Comparison between SC-I and IRVFL with various types of activation function }\label{fig:4}
\end{figure*}
\subsection{Discussions}
This section briefly explains why the performance of our proposed SC algorithms are better than MQ and IRVFL algorithms. Then, further remarks are provided  to highlight some characteristics of SCNs.

Although the universal approximation property can be ensured by maximizing an objective function (see Section 2), the Modified Quickprop (MQ) algorithm, aiming at iteratively finding the appropriate hidden parameters ($w$ and $b$) for the new hidden node, may face some obstacles in proceeding the optimization of certain objective function. In essence, it is a gradient-ascent algorithm with consideration of second-order information, and may be problematic when it is searching in a plateau over the error surface. That is to say, the well-trained parameters ($w$ and $b$) at certain iterative step fall into a region where both the first and second order of derivatives of the objective function are nearly zero, learning basically reaches a halt and can be terminated by a patience parameter for the purpose of time-saving. The overall speed for parameter updating might be quite slow. This issue seems severe for regression tasks. For implementations, MQ works less flexibly and seems very likely to fail in constructing a universal approximator, not to mention some inherent flaws including the selection of initial weights, learning rate, as well as a reasonable stopping criterion. Besides, there will be much computational burden in dealing with large-scale problems. The unique feature of the proposed SCNs is the use of randomness in learner model design with a supervisory mechanisms. Compared against Modified Quickprop algorithm, SC-II and SC-III are computationally manageable  and easy to be implemented. 

In addition, the output weights $\beta_L$ in IRVFL are analytically calculated by $e_{L-1}(X)^T\cdot h^{*}_L/(h_L^{*T}\cdot h^{*}_L)$ and then remain fixed in further phases. This constructive approach for computing the output weights, together with the freely random assignment of the hidden parameters, may cause very slow decreasing rate  of the residual error sequence, and as a result the learner model may fail to approximate the target function. Although this evaluation method for the output weights is also applied for SC-I, the whole construction process works successfully in building a universal approximator. In the end, some features of our proposed SCNs are summarized as follows:
\begin{itemize}
  \item The selection of $r$ that to some extent is directly associated with the error decreasing speed is quite important. In our experiments, its setting is unfixed and based on an increasing sequence stating from 0.9 and approaching 1. That makes it possible for the implementation of randomly searching $w_L$ and $b_L$. As the constructive process proceeds, the current residual error becomes smaller which makes the  configuration task on  $w_L$ and $b_L$ be more challenging (i.e., need more searching attempts as $L$ becomes larger). From our experience, this difficulty can be effectively alleviated by setting $r$ extremely close to 1.

  \item In SC algorithms, we used the scaling sigmoidal function in the hidden nodes, of which $\lambda$ may keep varying during the learning course. This implies that the scope setting for $w$ and $b$ should not be fixed. In our simulations, $\lambda$ is automatically decided from a given set $\{1,5,15,30,50,100,150,200\}$, in accompany with the setup of $r$. This strategy makes it possible for the built learner to possess multi-scale random basis functions rather than RVFL networks (as argued in \cite{SI-Gorban2016}) or IRVFL, and can inherently improve the probability of finding certain appropriate setting of the input weights and biases.

  \item In implementation,  $T_{max}$ controls the pool size of the random basis function candidates. It is related to both opportunity and efficiency, so we need to chose this parameter carefully with trade-off mind.  This operation aims at finding the most appropriate pairs of $w$ and $b$ that returns the largest $\xi_L$ (see Algorithm SC-I). Roughly speaking, this manipulation can be viewed as an alternative  to find a `suboptimal' solution of the maximization problem discussed in \cite{Kwok1997}.

  \item Based on the theoretical analysis given in Section 3, many  types of activation functions can be used in SCNs,  such as gaussian, sine, cosine and tanh function. Figure 4 shows both the training and test performance of SC-I and IRVFL with different choices of activation functions. It can be clearly seen that our SC-I algorithm is more efficient and effective than IRVFL as the random assignment of $w$ and $b$ from [-1,1] is unworkable and useless, no matter what kind of activation function is employed in the hidden layer.

 \item As a compromise between SC-I and SC-III, SC-II provides more flexibility in dealing with large-scale data modelling tasks, by right of its distinctive window-based recalculation for the output weights. Through optimizing a part of the output weights according to a given window size, SC-II not only outperforms SC-I, because the evaluation manner of the output weights in SC-I is not based on any objectives and no further adjustments are provided throughout the construction process; but also has some advantages over SC-III in building universal approximators for coping with large-scale data analytics. In practice, when the number of training samples is extremely large and $L>K$, the computation burden for calculating $H^{\dagger}_K$ in SC-II is far less than calculating $H^{\dagger}_L$ in SC-III. Besides, the robustness of SC-II with regard to the window size is favourable once $K$ is assigned from a reasonable range, as shown in Table 4 and 5. It is meaningful and interesting to find out more characteristics about SC-II in the future.
\end{itemize}

\section{Conclusions}
Our proposed framework in this paper provides with an alternative solution for building randomized learner models with supervisory mechanisms. As a powerful tool for fast data modelling, SCNs can be incrementally built by stochastically configuring the input weights and biases of each hidden node, and determining the output weights via constructive evaluation or solving an optimization problem for linear models. Each SCN model can be regarded as a specific  implementation of our SC  algorithms, which in theory ensure the convergence provided that newly generated nodes are kept adding to the model. From our hands-on experiences as demonstrated in the simulations,  SC-III exhibits the best performance in terms of the convergence rate, SC-I converges slowest but still outperforms IRVFL although they evaluate the output weights in the same way. It should be pointed that SC-II embedded with a given window size for updating the output weights makes a good trade-off between the efficiency and the scalability (i.e., for large-scale data analytics). 

In this work, we focus on SLFNN architecture with a sigmoidal basis/activation function for the hidden nodes. Some immediate  extensions to RVFL architecture  (with direct links between the inputs and the outputs) and Echo State Networks (with recurrent feedback at the hidden layer)  can be done easily. As a technical contribution to the machine learning community, our proposed stochastic configuration idea in this paper is significant and applicable for data representation with deep learning \cite{Graves2013,Hinton2006,Hinton2006science,LeCun2015}. Plenty of researches on SCNs and its applications can be explored in the future, and some of them have been done by our group. For instance, deep SCNs (DSCNs) and robust SCNs (RSCNs) have been developed for dealing with data representation and regression, and uncertain data analytics respectively, which  will be reported in our series of publications. Also, we are extending the present algorithms to ensemble learning, online learning and distributed learning. Except for these studies mentioned above, the following researches are being expected: robustness analyses of the modelling performance with respect to some key parameters involved in design of SCNs; and a guideline for selecting the basis/activation function so  that the modelling performance can be maximized.

%\textbf{Acknowledgment}

%The authors are grateful to  

\bibliographystyle{elsarticle-num}

\begin{thebibliography}{}
\small

\bibitem{Barron1993}
A. R. Barron, Universal approximation bounds for superpositions of a sigmoidal function, IEEE Transactions on Information Theory 39 (3) (1993) 930-945.

\bibitem{Broomhead1988}
D. S. Broomhead, D. Lowe, Multivariable functional interpolation and adaptive networks, Complex Systems, 2 (1988) 321-355.

\bibitem{Chen1995}
T. Chen, H. Chen, Universal approximation to nonlinear operators by neural networks with arbitrary activation functions and its application to dynamical systems, IEEE Transactions on Neural Networks 6(4) (1995) 911-917.

\bibitem{Cybenko1989}
G. Cybenko, Approximations by superpositions of a sigmoidal  function, Mathematics of Control, Signals and Systems 2 (1989) 303-314.

\bibitem{SI-Gorban2016}
A. N. Gorban, I. Y. Tyukin, D. V. Prokhorov, K. I. Sofeikov, Approximation with random bases: pro et contra, Information Sciences 364 (2016) 129-145.

\bibitem{Graves2013}
A. Graves, A. R. Mohamed, G. E. Hinton, Speech recognition with
deep recurrent neural networks, Proc. of IEEE International Conference
on Acoustics, Speech and Signal Processing (2013) 6645-6649. 

\bibitem{Hecht-Nielsen1988}
R. Hecht-Nielsen, Theory of the backpropagation neural network, Neural Networks  1(1) (1988) 593-605.

\bibitem{Hinton2006}
 G. E. Hinton, S. Osindero, and Y.-W. Teh, A fast learning algorithm
for deep belief nets, Neural Computation 18 (7) (2006) 1527-1554.

\bibitem{Hinton2006science}
G. E. Hinton, R. R. Salakhutdinov, Reducing the dimensionality
of data with neural networks, Science 313 (5786) (2006) 504-507.

\bibitem{Hornik1989}
K. Hornik, M. Stinchcombe, H. White, Multilayer feedforward networks are universal approximators, Neural Networks 1989, 2(5):359-366

%\bibitem{Hornik1990}
%K. Hornik, M. Stinchcombe, H. White, Universal approximation of an unknown mapping and its derivatives using multilayer feedforward networks, Neural %Networks 3 (5) (1990) 551-560.

\bibitem{Husmeier1999}
D. Husmeier, Random vector functional link (RVFL) networks. Neural Networks for Conditional Probability Estimation: Forecasting Beyond Point Predictions-Springer, (1999) Chapter 6.

\bibitem{Igelnik1995}
B. Igelnik, Y. H. Pao, Stochastic choice of basis functions in adaptive function approximation and the functional-link net, IEEE Transactions on Neural Networks 6 (6) (1995) 1320-1329.

\bibitem{Jones1992}
L. K. Jones, A simple lemma on greedy approximation in Hilbert space and convergence rates for projection pursuit regression and neural
network training, The Annals of Statistics 20 (1) (1992) 608-613.

\bibitem{Kwok1994}
T. Y. Kwok, D. Y. Yeung, Constructive neural networks: some practical considerations, Proc. of IEEE International Conference on Neural Networks, Orlando, Florida, USA, (1994) 198-203.

\bibitem{Kwok1997}
T. Y. Kwok, D. Y. Yeung, Objective functions for training new hidden units in constructive neural networks, IEEE Transactions on Neural Networks 8 (5) (1997) 1131-1148.

\bibitem{Lancaster1985}
P. Lancaster, M. Tismenetsky, The Theory of Matrices: With Applications, Academic Press, 1985.

\bibitem{LeCun2015}
Y. LeCun, Y. Bengio, G. E. Hinton, Deep learning, Nature 521 (2015) 436-444.

\bibitem{LiandWang2016}
M. Li, D. Wang, Insights into randomized algorithms for neural networks:
practical issues and common pitfalls, Information Sciences 382-383 (2017) 170-178.

\bibitem{LukoandJaeger2009}
M. Luko{\v{s}}Evi{\v{c}}Ius, H. Jaeger, Reservoir computing approaches
to recurrent neural network training, Computer Science Review 3(3) (2009) 127-149.

\bibitem{Mahoney2011}
M. W. Mahoney, Randomized algorithms for matrices and data, Foundations and Trends in Machine Learning 3 (2) (2011) 123-224.	

\bibitem{Pao1994}
 Y. H. Pao, G. H. Park, D. J. Sobajic, Learning and generalization characteristics of the random vector functional-link net, Neurocomputing 6 (2) (1994)
163-180.

\bibitem{Pao1992}
Y. H. Pao, Y. Takefji, Functional-link net computing: theory, system architecture, and functionalities, IEEE Computer 25 (5) (1992) 76-79.

\bibitem{Park1991}
J. Park, I. Sandberg, Universal approximation using radial basis function networks, Neural Computation 3 (1991) 246-257.

\bibitem{Rumelhart1986}
D. E. Rumelhart, G. E. Hinton, R. J. Williams, Learning internal representations by backpropagating errors, Nature 323 (1986) 533-536.

\bibitem{ScardapaneandWang2017}
S. Scardapane, D. Wang, Randomness in neural networks: An overview, WIREs Data Mining and Knowledge Discovery (2017) e1200. doi: 10.1002/widm.1200.

\bibitem{Schmidt1992}
W. F. Schmidt, M. A. Kraaijveld, R. P. Duin, Feedforward neural networks with random weights, Proc. of 11th IAPR International Conference on Pattern Recognition, Vol. II. Conference B: Pattern Recognition Methodology and Systems (1992) 1-4.

\bibitem{Tyukin2009}
I. Tyukin, D. Prokhorov, Feasibility of random basis function approximators for modeling and control, Proc. of IEEE Multi-Conference on Systems and Control, Saint Petersburg, Russia, (2009) 1391-1396.

\bibitem{SI-ProfWang2016}
D. Wang, Editorial: Randomized algorithms for training neural networks, Information Sciences 364-365 (2016) 126-128.


\end{thebibliography}

\end{document}